\documentclass[10pt,journal]{IEEEtran}

\usepackage{cite}
\usepackage{color}
\usepackage{algorithm}
\usepackage{algorithmic}
\usepackage{amsmath}
\usepackage{amssymb}
\usepackage{multirow} 
\usepackage{subfig}
\usepackage{amsthm}
\usepackage{graphicx}

\usepackage{url}

\usepackage[utf8x]{inputenc} 
\begin{document}

\title{Structured Graph Learning for Scalable Subspace Clustering: From Single-view to Multi-view}
%\author{Xiao Lu, Zhao~Kang, Xinwang Liu, Yang Yang,
\author{Zhao Kang, Zhiping Lin, Xiaofeng Zhu, Wenbo Xu
\IEEEcompsocitemizethanks{\IEEEcompsocthanksitem Z.Kang, Z.Lin, X.Zhu are with the School of Computer Science and Engineering, University of Electronic Science and Technology of China, Chengdu, Sichuan, 611731. (e-mail: zkang@uestc.edu.cn; 201921080534@std.uestc.edu.cn; seanzhuxf@gmail.com).
%201921080534@std.uestc.edu.cn%xuwenbo@uestc.edu.cn
\IEEEcompsocthanksitem W.Xu is with the School of Resources and Environment, University of Electronic Science and Technology of China, Chengdu, Sichuan, 611731. (e-mail: xuwenbo@uestc.edu.cn).
%\IEEEcompsocthanksitem C. Peng is with College of Computer Science and Technology, Qingdao University, Qingdao, China.
%E-mail: {cpeng@qdu.edu.cn}
%\IEEEcompsocthanksitem Q. Cheng is with Institute of Biomedical Informatics and Department of Computer Science, University of Kentucky, Lexington, KY, 40506.
%E-mail: {qiang.cheng@uky.edu}
%\IEEEcompsocthanksitem F. Nie is with the School of Computer Science, Northwestern Polytechnical University, Xi'an 710072, China, and also with the Center for OPTical IMagery Analysis and Learning, Northwestern Polytechnical University, Xi'an 710072, China.
%E-mail: { feipingnie@gmail.com}
%\IEEEcompsocthanksitem X. Liu is with School of Computer Science, National University of Defense Technology, Changsha 410073, China.
%E-mail: {xinwangliu@nudt.edu.cn}
%\IEEEcompsocthanksitem X. Peng is with College of Computer Science, Sichuan Univerisity, China.
}
%E-mail: {pengx.gm@gmail.com}}\\
\thanks{Manuscript received April 19, 2005; revised August 26, 2015.}}

\markboth{Journal of \LaTeX\ Class Files,~Vol.~14, No.~8, August~2015}%
{Shell \MakeLowercase{\textit{et al.}}: Bare Demo of IEEEtran.cls for Computer Society Journals}

\IEEEtitleabstractindextext{%
\begin{abstract}
%Graph-based methods have been widely applied in machine learning, pattern recognition, and data mining.
Graph-based subspace clustering methods have exhibited promising performance. However, they still suffer some of these drawbacks: encounter the expensive time overhead, fail in exploring the explicit clusters, and cannot generalize to unseen data points. In this work, we propose a scalable graph learning framework, seeking to address the above three challenges simultaneously. Specifically, it is based on the ideas of anchor points and bipartite graph. Rather than building a $n\times n$ graph, where $n$ is the number of samples, we construct a bipartite graph to depict the relationship between samples and anchor points. Meanwhile, a connectivity constraint is employed to ensure that the connected components indicate clusters directly. We further establish the connection between our method and the K-means
clustering. Moreover, a model to process multi-view data is also proposed, which is linear scaled with respect to $n$. Extensive experiments demonstrate the efficiency and effectiveness of our approach with respect to many state-of-the-art clustering methods.
  %In the field of clustering, subspace clustering has always been our research hotspot.We hope to get high accuracy through lower time complexity. Nowadays, many subspace clustering algorithms have been proposed, but the accuracy and time complexity can not reach a high level at the same time. Especially for the time complexity of big data, it has been a problem for us.In order to solve this problem,we proposed a novel structured graph subspace clustering algorithm in linear complexity.Inspired by bipartite graph with specified number of connected components and anchor graph,we construct a similarity matrix,which is a bipartite graph connected with raw data and anchor points.By limiting the rank of the Laplacian matrix,we divide the similarity matrix into k connected components,which is similar with k-means algorithm.And the optimization strategy adopts the method of updating parameters alternately and we can implement $k$-means on the final matrix to achieve the clustering results.Besides,our method can be extended to apply to the multi-view data and is able to solve out-of-sample problem.Extensive experiments on small-scale data,large scale-data,multi-view data and out-sample problem have proved our method's efficiency and effectiveness compared with many existing algorithm. 
\end{abstract}

\begin{IEEEkeywords}
Subspace clustering, multi-view learning, bipartite graph, connectivity constraint, out-of-sample, large-scale data, anchor graph.
\end{IEEEkeywords}}

\maketitle

\IEEEdisplaynontitleabstractindextext

\IEEEpeerreviewmaketitle

\vspace{1cm}
\IEEEraisesectionheading{\section{Introduction}\label{sec:introduction}}

\IEEEPARstart{A}{s} an unsupervised technique, clustering has always been an important research topic in machine learning, pattern recognition, and data mining. During the past few decades, a plethora of clustering methods have been developed, such as K-means \cite{jain2010data}, spectral clustering \cite{Ng2001On}, hierarchical clustering \cite{johnson1967hierarchical}, DBSCAN \cite{ester1996density}, deep clustering \cite{peng2019deep}, to name a few. %Weenp e make the similar objects are able to be classified in the same cluster.Many researchers proposed clustering methods over the past few decades
Recent years, to tackle the curse of dimensionality, subspace clustering has received increasing attention. It is capable of finding relevant dimensions spanning a subspace for each cluster \cite{vidal2011subspace,peng2018structured}. Among multiple approaches, graph-based subspace clustering often generates the best performance \cite{liu2019robust,li2017structured}. As a result, graph-based subspace clustering methods are in hot pursuit in recent years.%The principle of subspace clustering is based on the assumption that data can be expressed linearly as a series of subspace combinations.Based on this, many researchers have improved on this basis.

Specifically, some representative methods of this category include sparse subspace clustering (SSC) \cite{elhamifar2013sparse}, low-rank representation (LRR) \cite{liu2013robust}, least squares regression (LSR) \cite{lu2012robust}. To enjoy the benefit of discriminative features brought by deep neural works, some deep subspace clustering networks have recently been proposed \cite{ji2017deep,zhang2019self}. In general, they are implemented in two individual steps. Firstly, a $n\times n$ graph that represents the pairwise similarity between samples is learned. Secondly, the learned graph is input to spectral clustering algorithm which typically involves eigen-decomposition of the Laplacian matrix. This pipeline procedure has one flaw, i.e., the graph might potentially not be optimal for the downstream clustering since it might fail to achieve the cluster structure with exact cluster number \cite{zhu2019PR,kang2017twin}.

In particular, it is always difficult to capture complex similarity patterns for data with high-level semanticity (human level), e.g., speech, textual data, images, and videos \cite{lecun2015deep,kang2019robust}. Added to that, graph requires $\mathcal{O}(n^2)$ memory and eigen-decomposition often consumes $\mathcal{O}(n^3)$ time \cite{ren2020consensus}. Consequently, they are costly and make the processing of large-scale data prohibitive. Another usually ignored aspect is out-of-sample problem \cite{nie2011spectral}, i.e., the existing framework does not generalize to unseen data points. Putting it differently, for testing data points that are never seen during training, existing graph-based subspace clustering model cannot handle them because the ``graph structure" must be learned for all the data points during training.

Some recent research endeavors are devoted to reducing the algorithm complexity. For example, some off-the-shelf projection and sampling methods are applied for spectral clustering \cite{fowlkes2004spectral}. Bipartite graph is also widely used to speed up spectral clustering \cite{nie2017learning,chen2011large,chen2019labin}. To get rid of the ad-hoc functions used to calculate similarity, several scalable subspace clustering methods are proposed. Combining sparse representation and bipartite graph, Adler et al. \cite{adler2015linear}  propose a linear subspace clustering algorithm. \cite{wang2014exact} applies a data selection method to speedup computation. \cite{li2017large} adopts sampling and fast regression coding to cluster the codes of data points. \cite{You2016Scalable} employs orthogonal matching pursuit to reduce the computation load, but it often loses clustering accuracy. \cite{pourkamali2020efficient} proposes an efficient solver for sparse subspace clustering. Accelerated low-rank representation is also developed \cite{fan2018accelerated,xiao2015falrr}. \cite{peng2015unified} solves the large-scale challenge by converting it to out-of-sample problem. Though these techniques can alleviate the computation overhead, they often ignore the graph structure or fail to address out-of-sample problem.

Moreover, by virtue of the development of data acquisition and processing technologies, increasing volume of data are represented by multiple views \cite{huang2019mvstream,zhou2019incremental,tao2019joint,TangTMM2018,zhuge2017unsupervised}. For example, a video might consist of text, images, and sounds \cite{yin2018multiview,sun2019multi}; an image can be described in different features, e.g., SIFT, GIST, LBP, HoG, and Garbor \cite{zhan2018graph,chen2020multi}; a document can be translated into different languages \cite{yao2019multi,wang2019gmc}. These heterogeneous features often supply complementary information that could be helpful for our tasks at hand \cite{hou2017multi,wang2016iterative,wang2020survey}. As a result, multi-view subspace clustering has also been investigated \cite{zhang2018generalized,zhang2019robust,chen2019jointly}. For instance, Cao et al.\cite{cao2015diversity} consider both the consistency and the diversity among the multiple views; Gao et al.  \cite{gao2015multi} learn multiple graphs and let them correspond to a unique cluster indicator matrix; Zhang et al. \cite{zhang2018generalized,tao2020latent} perform subspace clustering in latent space; Kang et al. \cite{kang2020partition} propose to fuse multi-view information in partition space. These algorithms often achieve better results than the single view methods.

Unfortunately, most of existing multi-view subspace clustering methods also encounter scalability problem, which hinders their applications on large-scale data. Some single-view large-scale subspace clustering cannot be directly extended to cope with multi-view data. Recently, Kang et al. \cite{kang2019large} make the first effort to tackle large-scale multi-view subspace clustering challenge. However, it ignores the graph structure and fails to deal with unseen data.

In this paper, we simultaneously deal with the three issues of subspace clustering, i.e., high complexity, explicit graph structure and out-of-sample. Apart from single-view model, we also propose a multi-view model, seeking for a structured graph that compatibly crosses multiple views. Firstly, we build a dictionary matrix by selecting $m$ landmarks from raw data by K-means algorithm. Secondly, we learn the relationship between our raw data and the landmarks, which generates a bipartite graph with $k$ connected components if the data contains $k$ clusters. Thus, a cluster indicator matrix is naturally obtained. Thirdly, for multi-view data, extra view-wise weights are introduced to discriminate different views. The advantages of our approach are: the small affinity matrix can preserve the manifold of the data; the constraint of the bipartite graph discovers the underlying cluster structure; our method addresses out-of-sample problem; our overall complexity is linear to the size of data. Compared to state-of-the-art techniques, our methods gain a lot in terms of effectiveness and efficiency.

In a nutshell, the key contributions of this paper are:
\begin{itemize}
\item{We present a novel structured graph learning framework for large-scale subspace clustering in linear time. Besides, this method also solves the out-of-sample problem.}
\item{A scalable multi-view subspace clustering method is proposed. The bipartite graph, cluster indicator matrix, and view-wise weights learn from each other interactively and supervise each other adaptively.}
\item{Theoretical analysis establishes the connection between our method and K-means clustering. Extensive experimental results demonstrate the superiority of our method with respect to many state-of-the-art techniques. }
\end{itemize}

The rest of our paper is as follows. In Section \ref{back}, we introduce some background on subspace clustering and anchor graph. Section \ref{propose} presents our proposed graph learning model and solution. In Section \ref{theory}, we perform theoretical analysis and compare time complexity. Section \ref{multi} extends our method to the multi-view data. Single-view and multi-view experiments are conducted in Section \ref{singleexp} and \ref{multiexp}, respectively. This paper is rounded up with a conclusion in Section \ref{conclude}.

\section{Background}\label{back}
In this section, we present an overview of the single-view and multi-view subspace clustering, and then introduce the anchor graph.
\subsection{Subspace Clustering}
Given data $X\in\mathcal{R}^{d\times n}$, which includes $n$ samples each with $d$ features, subspace clustering assumes that each data sample can be expressed as a combination of other data points in the same subspace. This combination coefficient matrix is considered as the similarity graph, which captures the global structure of data. In general, the following model is solved \cite{kang2017twin},
\begin{equation}
\min_S \quad \|X-XS\|_F^2+\alpha f(S)\quad s.t.\quad S\geq 0,\hspace{0.1cm}  S\textbf{1}=\textbf{1},
\end{equation}
where $S\in\mathcal{R}^{n\times n}$ is the nonnegative similarity matrix and $\alpha>0$ is a balance parameter. The first term is the reconstruction error and the second term $f(\cdot)$ is a regularizer function, including low-rank constraint \cite{liu2013robust,kang2020structure}, sparse $\ell_1$ norm \cite{elhamifar2013sparse}, Frobenius norm \cite{kang2020relation,peng2016constructing}. $\textbf{1}$ is a vector of ones and $S\textbf{1}=\textbf{1}$ means that each row of $S$ adds up to 1. 

It can be seen that the size of graph $S$, which often results in $\mathcal{O}(n^3)$ computation complexity, would be a burden on both computation and storage for large-scale data. Furthermore, the subsequent spectral clustering step also suffers $\mathcal{O}(n^3)$ complexity. Some recent subspace clustering methods with low complexity have been developed. For instance, SSCOMP \cite{You2016Scalable} is built on orthogonal matching pursuit; ESSC \cite{pourkamali2020efficient} is a proximal gradient framework to solve sparse subspace clustering; ALRR \cite{fan2018accelerated} is a faster solver for low-rank representation. Though they can deal with large-scale data, they fail to cope with out-of-sample problem and ignore the graph structure. Kang et al. \cite{kang2017twin} consider that the graph should have exactly $k$ components if the data contain $k$ clusters. However, it has $\mathcal{O}(n^3)$ complexity and cannot address out-of-sample challenge. For new samples, SLSR \cite{peng2015unified} projects them into the union of subspaces spanned by in-sample data. However, SLSR does not explicitly consider the graph structure. In practice, the data can display structures beyond simply being low-rank or sparse \cite{haeffele2019structured}. In summary, there is no single method which can address all three challenges faced by subspace clustering: high complexity, graph structure, out-of-sample. 

For multi-view subspace clustering, more attention is paid to how to boost the clustering accuracy by fully exploring the complementary information carried by multi-view data. For instance, multi-view low-rank sparse subspace clustering (MLRSSC) \cite{brbic2018} harnessing both low-rank and sparsity constraints shows much better performance than previous methods, such as \cite{xia2014robust}. Multi-view subspace clustering with intactness-aware
similarity (MSC\_IAS) \cite{wang2019multi} tries to construct a graph in latent space, which leads to superior accuracy. Kang et al. \cite{kang2019large} make the first attempt to address the scalability issue of multi-view subspace clustering. Though it has a linear complexity, it fails to consider the discriminative nature of different views. Moreover, the graph learning and clustering stage are separated, so the graph structure is ignored.

\subsection{Anchor Graph}
Anchors or landmarks were previously used in scalable spectral clustering \cite{chen2011large}. Basically, its idea is to select a small set of data samples called anchors or landmarks to represent the neighborhood structure. Typically, these representative points are chosen based on K-means method (the cluster centers) or random sampling \cite{wang2016scalable}. Specifically, with $m$ anchors $A=\{a_1,\cdots,a_m\}\in\mathcal{R}^{d\times m}$, a small graph $Z\in\mathcal{R}^{n\times m}$ is built to measure the relationship between the anchors and the whole data. Typically, $Z$ is constructed by Gaussian kernel function \cite{chen2011large,han2017orthogonal}, which might not be flexible enough to characterize complex data.

\begin{figure}
    \centering
    \includegraphics[width=.45\textwidth]{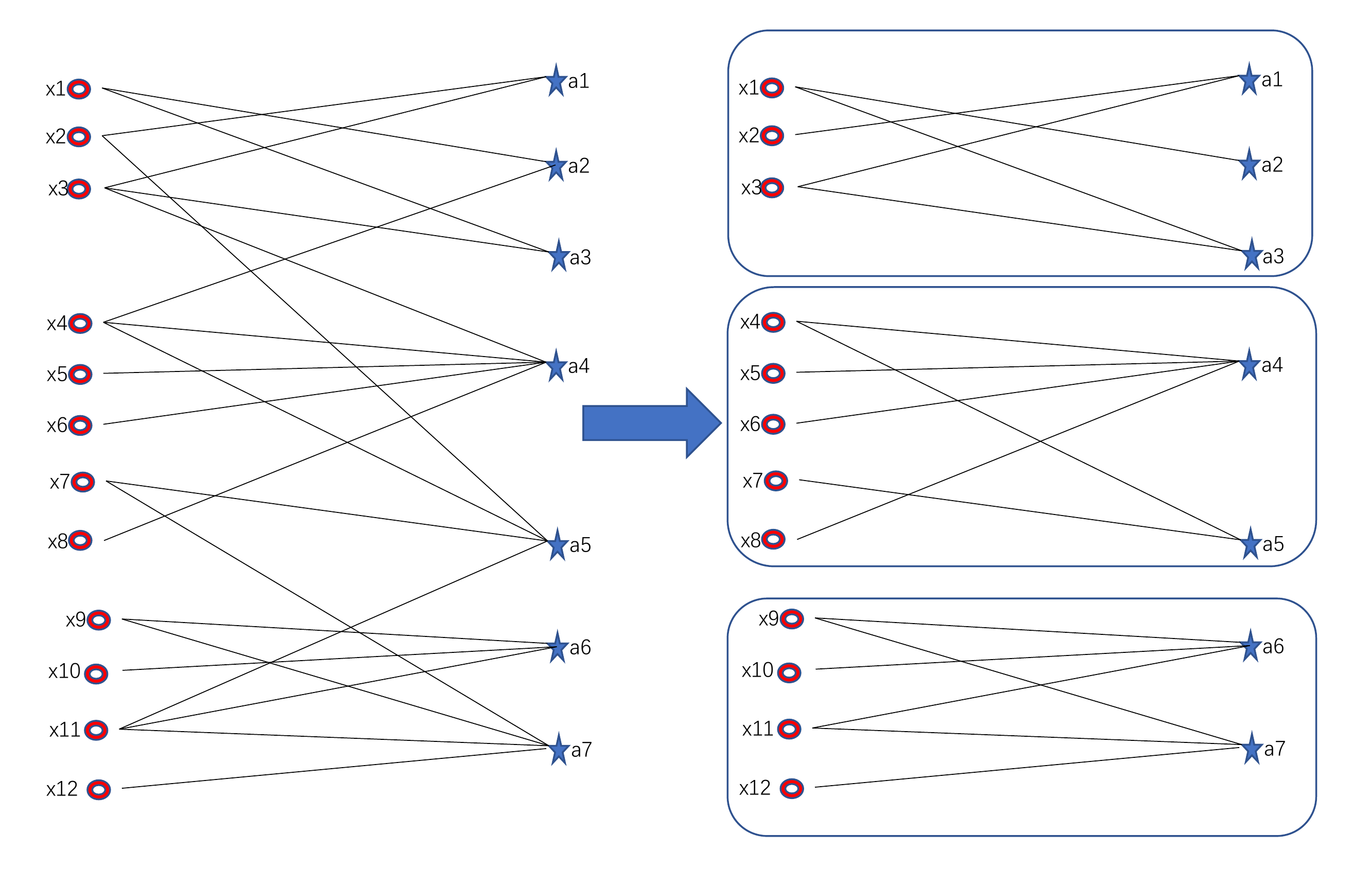}
    \caption{The optimal bipartite graph with constraint. Initially, the nodes in the left of the bipartite graph are connected randomly with the right anchors. After enforcing the constraint, the structured bipartite graph contains a specified number of connected components. }
    \label{bp}
\end{figure}

For subspace clustering, we can borrow the idea of anchor and treat $A$ as a dictionary \cite{kang2019large}. Afterwards, we can solve the following model to learn $Z$ automatically, i.e.,
%\begin{equation}
%Z_{ij}=
 %\begin{cases}\frac{K_\delta(x_i,a_j)}{\sum_{j'\in<i>}K_\delta(x_i,a_{j'})}, &j\in<i>\\
 %  0, &\text{otherwise}
 %   \end{cases}
 %   \label{hand}
%\end{equation}
%where $K_\delta (\cdot)$ is Gaussian kernel with paramenter $\delta$ and we are also able to use other kernels,and $<i>$ indicates the indexes of s nearest neighbours around $x_i$.Based on the idea of the anchor graph and consider the difficulty of choosing the hyperparameter  $\delta$,an efficient way is proposed to learn anchor graph from our raw data,then we should solve the following problem:
\begin{equation}
\begin{split}
\min_{Z}\quad \|X-AZ^\top\|_F^2+\alpha \|Z\|_F^2\\ s.t.\hspace{.15cm} 0\leq Z,\hspace{.1cm} Z\textbf{1}=\textbf{1}.
\end{split}
\label{sc}
\end{equation}
Though Eq. (\ref{sc}) is quite simple, it does not consider any cluster structure. Putting it differently, $Z$ might be just one connected component, as shown in the left side of Fig. \ref{bp}. It would be desired if it has exactly $k$ connected components denoted by $Z\in\Omega$ \cite{nie2017learning}, as shown in the right part of Fig. \ref{bp}. Then, problem (\ref{sc}) becomes
\begin{equation}
\begin{split}
\min_{Z}\quad  \|X-AZ^\top\|_F^2+\alpha \|Z\|_F^2\\ s.t.\hspace{.15cm} 0\leq Z,\hspace{.1cm} Z\textbf{1}=\textbf{1}, Z\in\Omega.
\end{split}
\label{sc2}
\end{equation}
In the next section, we will show how to address this challenging problem. 
%where the size of matrix Z is $n\times m$,which is constructed by learning form data instead of using equation(\ref{hand}) 
%\subsection{spectral clustering}
%If we obtain the Laplacian matrix L we can apply spectral clustering on the matrix to achieve the indicator matrix as follow\cite{Ng2001On}:
%\begin{equation}\min _{F^{T} F=I} \operatorname{Tr}\left(F^{T} L F\right)\end{equation}
%where matrix $F$ indicates the indicator matrix whose size is $n\times k$(k is the number of clusters) and we can conduct k-means on the indicator matrix to get the results. 
\section{Structured Graph Learning}\label{propose}
To explicitly explore the cluster structure of $Z$, we can make use of bipartite graph. To be precise, bipartite graph $S$ associated with $Z$ is defined by $S=\left[\begin{array}{cc}{} & {Z} \\ {Z^{\top}} & {}\end{array}\right]\in\mathcal{R}^{(n+m)\times (n+m)}$. Then, the normalized Laplacian matrix $L$ is expressed as $L=I-D^{-\frac{1}{2}} S D^{-\frac{1}{2}}$, where $D$ is a diagonal matrix with its $i$-th diagonal element defined as $d_{i}=\sum_{j=1}^{n+m} s_{i j}$. According to spectral graph theory \cite{chung1997spectral}, the normalized Laplacian has the following property:
%For large-scale data,by setting the bipartite graph's(anchor graph's) components to be the clusters on subspace clustering,we can establish an unified framework to train Z.

\newtheorem{thm}{\bf Theorem}%[section]
\begin{thm}\label{thm1}
The number of connected components in $S$ is equal to the cardinality $k$ of the 0 eigenvalue of $L$.
%The multiplicity $k$ of the eigenvalue 0 of the normalized Laplacian matrix $L$ is equal to the number of connected components in the bipartite graph associated with $Z$.
\end{thm}
By the theorem \ref{thm1}, if $\operatorname{rank}(L)=(n+m)-k$ meets, the $n$ data samples and $m$ anchors are grouped into $k$ clusters. Therefore, to achieve the ideal subspace clustering with specified $k$ clusters, we can explicitly express the constraint in problem (\ref{sc2}). It yields
\begin{equation}
\begin{split}
\min_{Z}\quad  &\|X-AZ^\top\|_F^2+\alpha \|Z\|_F^2\\
s.t.&\hspace{.15cm} 0\leq Z,\hspace{.1cm} Z\textbf{1}=\textbf{1}, \operatorname{rank}\left(L\right)=(n+m)-k.
\end{split}
\label{method1}
\end{equation}
%This equation add a constraint $\operatorname{rank}\left(L\right)=(n+m)-k$ and it means our bipartite graph has $k$ connected components\cite{chung1997spectral},which make our data divide into k clusters.
Considering that the rank constraint is hard to tackle, we can relax the constraint by following \cite{kang2020structured}. Eventually, our proposed Structured Graph Learning (SGL) framework for Subspace Clustering can be formulated as
%\begin{equation}
%\begin{split}
%\begin{aligned}
%\min_{Z} \|X-AZ^\top\|_F^2+\alpha \|Z\|_F^2+\beta \sum_{i=1}^{k} \sigma_{i}\left(L\right)\\
%s.t.\hspace{.15cm} 0\leq Z,\hspace{.1cm} Z\textbf{1}=\textbf{1}.
%\end{aligned}
%\end{split}
%\label{method2}
%\end{equation}
%where $\sigma_{i}\left(L\right)$ expresses the $i$-th smallest eigenvalue of $L$.By converting $\operatorname{rank}\left(L\right)=(n+m)-k$ into $\sigma_{i}\left(L\right)$,it almost has no difference when the value of $\beta$ is large enough.Because it will make the last term to be zero,thus making the constraints satisfied.
%Then we use the Ky Fan's theorem\cite{fan1949theorem}:
%\begin{equation}
%\sum_{i=1}^{k} \sigma_{i}\left(L\right)=\min _{F \in R^{(n+m) \times k}, F^{T} F=I} \operatorname{Tr}\left(F^TLF\right)
%\label{method4}
%\end{equation}  

%\begin{equation}
%\begin{split}
%\min_{Z} \|X-AZ\|_F^2+\alpha \|Z\|_F^2\\ s.t.\hspace{.15cm} 0\leq Z,\hspace{.1cm} Z\textbf{1}=\textbf{1}.
%\end{split}
%\label{sc}
%\end{equation}
%Finally we can rewrite our equation as:
\begin{equation}
\begin{split}
%\begin{aligned}
\min_{Z,F}&\quad \|X-AZ^\top\|_F^2+\alpha \|Z\|_F^2+\beta Tr(F^{\top}LF)\\
&s.t.\hspace{.15cm} 0\leq Z,\hspace{.1cm} Z\textbf{1}=\textbf{1}, F^{\top}F=I,
%\end{aligned}
\end{split}
\label{method11}
\end{equation}
%where $S=\left[\begin{array}{cc}{} & {Z} \\ {Z^{T}} & {}\end{array}\right]$ and the normalized Laplacian matrix $L$ is expressed as $L=I-D^{-\frac{1}{2}} S D^{-\frac{1}{2}}$.And D is a diagonal matrix whose size is $(n+m) \times(n+m)$ and its $i$-th diagonal element is $d_{i i}=\sum_{j} s_{i j}$.In this equation,we can update F and Z alternately.
where $F\in\mathcal{R}^{(n+m)\times k}$. It is worth pointing out that the above structured graph learning can also be applied to semi-supervised classification \cite{bo2019latent}. This problem can be solved by an alternating optimization strategy.
\subsection{Optimization Strategy}
We solve $Z$ and $F$ iteratively, i.e., fix one of them and then update the other one.
\subsubsection{Fix $F$ and Solve $Z$}note that $L$ is $I-D^{-\frac{1}{2}} S D^{-\frac{1}{2}}$, so both $S$ and $D$ depend on variable $Z$. Fortunately, we can employ the following equation
\begin{equation}
\operatorname{Tr}\left(F^{\top}LF\right)=\frac{1}{2} \sum_{i=1}^{(n+m)} \sum_{j=1}^{(n+m)} s_{i j}\left\|\frac{F_{i,:}}{\sqrt{d_{i}}}-\frac{F_{j,:}}{\sqrt{d_{j}}}\right\|_{2}^{2}
\end{equation}
Considering the structure of $S$, we can further convert above formula into the following formulation
\begin{equation}
\operatorname{Tr}\left(F^{\top}LF\right)=\sum_{i=1}^{n} \sum_{j=1}^{m} z_{i j}\left\|\frac{F_{i,:}}{\sqrt{d_{i}}}-\frac{F_{n+j,:}}{\sqrt{d_{n+j}}}\right\|_{2}^{2}
\end{equation}
Defining $\left\|\frac{F_{i,:}}{\sqrt{d_{i}}}-\frac{F_{n+j,:}}{\sqrt{d_{n+j}}}\right\|_{2}^{2}$ as $W_{ij}$, we can solve our problem row by row as
\begin{equation}
\begin{split}
%\begin{aligned}
\min_{Z_{i,:}}&\quad Z_{i,:}A^{\top}AZ_{i,:}^T-2X_{:,i}^{\top}AZ_{i,:}^{\top}+\alpha Z_{i,:}Z_{i,:}^{\top}+\beta W_{i,:}Z_{i,:}^{\top}\\
& s.t.\hspace{.15cm} 0\leq Z_{ij}\leq 1, \sum_{j}Z_{ij} =1.
%\end{aligned}
\end{split}
\label{solz}
%\label{method}
\end{equation}

This problem can be easily solved via convex quadratic programming.% and  get the Z. Fig. (\ref{bp}) demonstrates the learned $Z$ which has exactly $k$ connected components.
\subsubsection{Fix $Z$ and Solve $F$}
when Z is fixed, the first and the second term in Eq. (\ref{method11}) become constant. Our problem can be equivalently written as
\begin{equation}
\max _{F \in \mathbb{R}^{(n+m) \times k}, F^{\top} F=I} \operatorname{Tr}\left(F^{\top} D^{-\frac{1}{2}}S D^{-\frac{1}{2}} F\right)
\label{solfo}
\end{equation}
In general, computing the eigenvectors of $S$ takes $\mathcal{O}(k(n+m)^2)$. To circumvent this complexity, we employ the special structure of $S$ and compute the eigenvectors of $Z$ instead. Concretely, we decompose $F$ and $D$ as 
\begin{equation}
F=\left[\begin{array}{l}
{U} \\
{V}
\end{array}\right], D=\left[\begin{array}{cc}
{D_{U}} & {} \\
{} & {D_{V}}
\end{array}\right],
\end{equation}
where $U \in \mathcal{R}^{n \times k}, V \in \mathcal{R}^{m \times k}, D_{U} \in \mathcal{R}^{n \times n}, D_{V} \in \mathcal{R}^{m \times m}$, Eq.(\ref{solfo}) can be rewritten as following
\begin{equation}
\max _{U^{\top} U+V^{\top} V=I} \operatorname{Tr}\left(U^{\top} D_{U}^{-\frac{1}{2}} Z D_{V}^{-\frac{1}{2}} V\right).
\label{solf}
\end{equation}
According to the following theorem \cite{nie2017learning}, it can be easily solved.
\begin{thm}\label{thm2}
Suppose $Q \in \mathcal{R}^{n \times m}, X \in \mathcal{R}^{n \times k}, Y \in \mathcal{R}^{m \times k} .$ The optimal
solutions to the problem
$$
\max _{X^{\top} X+Y^{\top} Y=I} \operatorname{Tr}\left(X^{\top} Q Y\right)
$$
are $X=\frac{\sqrt{2}}{2} U_{1}, Y=\frac{\sqrt{2}}{2} V_{1},$ where $U_{1}$ and $V_{1}$ are corresponding to the top $k$ left and right singular vectors of $Q$, respectively.
\end{thm}
The complete algorithm for subspace clustering is summarized in Algorithm 1. 
By the theory of alternating optimization \cite{bezdek2003convergence}, the objective function
value of our problem (\ref{method11}) will monotonically decrease in each iteration. Moreover, since the
objective function has a lower bound, such as zero, the above iteration converges.
\subsection{Out-of-sample Problem}\label{outof}
Out-of-sample problem is hard and few discussed for subspace clustering. This is because we must compute the graph consists of all data points, which is inherently impossible to just involve unseen data. In contrast, our SGL method can handle new data. Note that our method also outputs embedding vectors $V$ and cluster labels for the anchor points. Then, we just need to implement the classic k-Nearest Neighbor (kNN) algorithm, which will propagate the labels to the new data. For each new data point, this process takes $\mathcal{O}(md)$ time, which is far lower than the cost carried out on the training data $\mathcal{O}(nd)$. Therefore, our method can handle the out-of-sample problem efficiently.

\section{Theoretical Analysis}\label{theory}
In this section, we demonstrate that our method is connected to K-means method and provide the computational complexity analysis.
\subsection{Relationship with K-means Algorithm}
%We can observe that when we select some certain parameter our method is equal to K-means algorithm.

\begin{thm}\label{thm4}
When we let $\alpha$ go to $\infty$, our problem (\ref{method1}) is equal to K-means problem.  
\end{thm}
\begin{proof}\label{proof 4}
In Eq. (\ref{method1}), we set a constraint to make $Z$ satisfy this property: each component contains several data points and anchor points; the number of connected components is $k$, which means all the points are classified into clusters. Thus the value of $\beta\operatorname{Tr}\left(F^{\top}LF\right)$ would be equal to zero. Denote the $i$-th component of $Z$ by $Z^{i} \in \mathcal{R}^{m_{i} \times n_{i}}$, where $n_{i}$ represents the data points number corresponding to this component and $m_{i}$ denotes the anchor number corresponding to this component. Hence, solving Eq. (\ref{method11}) is to solve the following problem for each component of $Z$
\begin{equation}
\begin{aligned}
&\min _{Z^{i}}\left\|X^{i}-A^{i} (Z^{i})^{\top}\right\|_{F}^{2}+\alpha\left\|Z^{i}\right\|_{F}^{2}\\
&\text { s.t. } \quad Z^{i} \mathbf{1}=\mathbf{1}, \quad 0 \leq Z^{i} \leq 1,
\end{aligned}
\end{equation}
where $X^{i}$ and $A^{i}$ consist of samples and anchors corresponding to the $i$-th component of $Z$. When $\alpha \rightarrow \infty,$ the above problem becomes:
\begin{equation}
\begin{array}{c}
\min\limits_{Z^{i}}\quad \left\|Z^{i}\right\|_{F}^{2} \\
\text {s.t.}\quad Z^{i} \mathbf{1}=\mathbf{1}, 0 \leq Z^{i} \leq 1
\end{array}
\end{equation}
The optimal solution is that all elements of $Z^{i}$ are equal to
$\frac{1}{m_{i}}$.
Thus, when $\alpha \rightarrow \infty$, the optimal solution $Z$ to problem (\ref{method1}) is:
\begin{equation}
z_{i j}=\left\{\begin{array}{ll}
\frac{1}{m_{p}},&x_{i}\text{ and }a_{j}\text{ in the same $p$-th} \text { component } \\
0, & \text { otherwise}
\end{array}\right.
\end{equation}
Let's denote the solution set of this partition as $\mathcal{K}$. It can be shown that $\|Z\|_{F}^{2}=k$. Thus Eq.(\ref{method1}) becomes
\begin{equation}
\min_{Z_i \in \mathcal{K}}\left\|X_i-AZ_i^{\top}\right\|^{2}.
\label{km}
\end{equation}
We can find that Eq. (\ref{km}) is exactly the objective function in K-means method and $AZ_i^{\top}$ denotes the centroid of cluster $i$. Therefore, the problem (\ref{method1}) is the problem of K-means. 
\end{proof}
% \begin{table*}			
% 	\renewcommand{\arraystretch}{1.1}
	
% 	\setlength{\tabcolsep}{2pt}{
% 		\begin{center}
% 			\caption{Computational complexity,where n is number of data points and d is size of the features.m is the number of pre-defined upper bound for the rank of the coefficients matrix.$t$ an $t_1$ is the number of iterations.$m_1$ is the number of anchors.k is the number of the clusters.$k_m$ and $m_2$ is a parameter.
% \label{comp}}
% 		\begin{tabular}{|ccccccc}  
% 		\hline  % 表格的横线
% 		&ALRR &KMM &ESSC &FNC &SSCOMP&our method\\[-6pt]  %可以避免文字偏上来调整文字与上边界的距离
% 		& &  & & & &\\[-6pt]  %可以避免文字偏上 
% 		\hline
% 		& &  & & & &\\[-6pt] 
%         $\mathcal{O}$&$\mathcal{O}(4dmn)$ & $O\left(n\left(\left(m d+m c+m^{2}\right) t_{1}+m d\right) t\right)$ &$\mathcal{O}(n(\log n))$ &$\mathcal{O}(nmd+nmk)$&$\mathcal{O}(n(n-1)k_m+(m_2)^3+nk^2t)$ &$\mathcal{O}(nm_1^3t+m_1nt+nm_1t_1d+nk^2t_1)$\\[-6pt] %可以避免文字偏上 
% 		& & & & & \\
% 		\hline
% 		\end{tabular}

% 	\end{center}		}
	
% \end{table*}
\begin{table}[H]			
			\renewcommand{\arraystretch}{1.5}
			\setlength{\tabcolsep}{12pt}{
			\begin{center}
			\caption{Summary of computational complexity of various methods. In this table, $n$ is number of data points and $d$ is size of features. $m_1$ is the number of pre-defined upper bound for the rank of the coefficient matrix. $t$ is the number of iterations until the convergence. $t_1$ is number of iterations for the K-means algorithm. $t_2$ is the number of iterations of the sub-alternating system. $m$ is the number of anchors. $k$ is the number of the clusters.\label{comp}}
				\begin{tabular}{|c|c| }
					\hline
					 Method&Time complexity \\
					\hline	
			ALRR&$\mathcal{O}(4dm_1n)$\\\hline
			KMM&$\mathcal{O}\left(n\left(\left(m d+m c+m^{2}\right) t_{2}+m d\right) t\right)$\\\hline
			ESSC&$\mathcal{O}(n\log n)$\\\hline
			FNC&$\mathcal{O}(nmd+nmk)$\\\hline
			SSCOMP&$\mathcal{O}(n^2dt)$\\\hline
			SGL&$\mathcal{O}(nm^3t+2mnt+nmt_1d+nk^2t_1)$\\

					\hline	
			\end{tabular}
	\end{center}		}
	\end{table}
%\subsubsection{Relationship with spectral clustering}
%in spectral clustering, we have to input affinity matrix $Z$ before we apply spectral clustering on the problem:
%\begin{equation}\min _{F^{T} F=I} \operatorname{Tr}\left(F^{T} L F\right)\end{equation}
%The $k$ smallest eigenvalues of matrix $L$ makes up the indicator matrix $F$.In general,the premise of using $F$ for clustering is that $Z$ has totally $k$ connected components.After obtaining F,we can do k-means or other commonly used clustering methods on it to get the final clustering result.

%Different from spectral clustering,we obtain $Z$ by training.In other words,our similarity matrix $Z$ is not predefined.We optimize $Z$ in every iteration,which means we consider the effect of clustering on $Z$.Specifically,we require our similarity matrix $Z$ should be divided into $k$ connected components by using matrix $F$.Correspondingly,matrix $L$'s k smallest matrix make up the matrix $F$.Our method's advantage compared with those spectral methods is that we optimize matrix $Z$ and matrix $F$ simultaneously,while other methods assign matrix Z.In every iteration,we optimize matrix $Z$ and $F$ alternately.Finally,we will be able to obtain a significant results compared with those existing spectral methods.
\begin{algorithm}[tb]
\caption{SGL for Subspace Clustering}
\label{alg:algorithm}
\textbf{Input}: Data matrix $X\in\mathcal{R}^{d\times n}$, anchor matrix $A\in\mathcal{R}^{d\times m }$, cluster number $k$, parameters $\alpha$ and $\beta$ \\
\textbf{Output}: $k$ clusters
\begin{algorithmic}[1] %[1] enables line numbers
\STATE Initialize the matrix $F$ randomly.
%\STATE Apply K-means algorithm on data matrix $X$ to obtain landmark matrix $A$.
\WHILE{convergence condition does not meet}
\STATE Update $Z$ in Eq. (\ref{solz}) via convex quadratic programming
\STATE Update $U$ in Eq. (\ref{solf}) by Theorem \ref{thm2}
\ENDWHILE
\STATE Run K-means on matrix $U$ to achieve the final partition
%\STATE \textbf{return} matrix $U$ of matrix $F$
\end{algorithmic}
\end{algorithm}

\subsection{Complexity Analysis}
The proposed method uses the anchor idea to construct a smaller graph $Z\in\mathcal{R}^{n\times m} (m\ll n)$ and performs singular value decomposition (SVD) on a smaller matrix $Z$, so that the complexity can be reduced significantly. Specifically speaking, denote $t$ as the iteration number, we implement SVD on $Z$ in each iteration to obtain indicator matrix $U$, which takes $\mathcal{O}(m^3t+mnt)$. The $W$ computation costs $\mathcal{O}(mnt)$. We apply the built-in Matlab function $quadprog$ to solve $Z$ in each iteration, which leads to $\mathcal{O}(nm^3t)$. As shown in Eq. (\ref{solz}), $Z$ can be efficiently solved in parallel. Besides, we need extra $\mathcal{O}(nmt_1d)$ time to build the dictionary $A$ by K-means algorithm and $\mathcal{O}(nk^2t_1)$ time to achieve the final results by applying K-means on $U$, where $t_1$ is another iteration number. In conclusion, the time complexity of our method is linear to the number of points $n$.%$\mathcal{O}(n)$.

We compare the time complexity of several recent scalable clustering methods in Table \ref{comp}. As we can see, most methods have a linear complexity. As shown in experimental part, some of them sacrifice accuracy in pursuit of improving time efficiency.

\section{Multi-view Structured Graph Learning}\label{multi}
 Compared to single-view scenario, large-scale multi-view subspace clustering is few studied. In this section, we show that our structured graph learning method can be easily extended to handle multi-view data $[X^1, X^2, \cdots, X^c]$, where $X^v\in\mathcal{R}^{d^{(v)}\times n}$ is the $v$-th view data matrix with $d^{(v)}$ features. For multi-view clustering, all views are required to share the same cluster pattern. Therefore, it is reasonable to assume that there exists a unique graph $Z$. However, different views might play various roles, so we introduce a weight factor $\lambda^v$ to balance the importance of different views. Finally, our proposed Multi-view Structured Graph Learning (MSGL) method can be formulated as
\begin{equation}
\begin{split}
%\begin{aligned}
\min_{Z,F,\{\lambda^v \geq 0\}}&\sum\limits_{v=1}^c \lambda^v\|X^v-A^vZ^\top\|_F^2+\alpha \|Z\|_F^2+\beta Tr(F^{\top}LF)\\+\sum\limits_{v=1}^c (\lambda^v)^\gamma
  &\hspace{.2cm}s.t.\hspace{.2cm} 0\leq Z,\hspace{.05cm} Z\textbf{1}=\textbf{1}, F^{\top}F=I,
%\end{aligned}
\end{split}
\label{multiview}
\end{equation}
where $\gamma<0$. In this model, different anchor points are generated for different views. %the anchor points are also adopted by k-means through all the views.We add a parameter $\lambda^v$ to integrate all the views to a whole model. 
In a similar way as SGL, we can optimize the three variables alternatively.
%\subsection{Optimization Strategy}
\subsection{Fix $\lambda^v$, $F$, Update $Z$}
%When we fix $\lambda^v$ and F,as we proved in section 2:
The sub-problem we are going to solve is
%\begin{equation}
%\operatorname{Tr}\left(F^{T}LF\right)=\sum_{i=1}^{n} \sum_{j=1}^{m} z_{i j}\left\|\frac{f_{i}}{\sqrt{d_{i}}}-\frac{f_{n+j}}{\sqrt{d_{n+j}}}\right\|_{2}^{2}
%\end{equation}
%Besides,the last term of the equation we add is regarded as constant.This problem is equal to a convex quadratic programming problem as follow and can be easily solved by the built-in function in Matlab.
\begin{equation}
\begin{split}
%\begin{aligned}
&\min_{Z_{i,:}} \sum_{v=1}^{c}\lambda^{v}(Z_{i,:}(A^v)^{\top}A^vZ_{i,:}^{\top}-2(X^v_{:,i})^{\top}A^vZ_{i,:}^{\top})+\alpha Z_{i,:}Z_{i,:}^{\top}\\
&+\beta W_{i,:}Z_{i,:}^{\top}
\hspace{.2cm}  s.t. \hspace{.2cm} 0\leq Z_{ij}\leq 1, \sum_{j}Z_{ij} =1.
%\end{aligned}
\end{split}
\label{solmz}
\end{equation}
This quadratic problem can be easily solved.
\subsection{Fix $\lambda^v$, $Z$, Update $F$}
%The last term of the equation is also a constant.The problem is also equal to that in section 2.Our final problem is written as follow:
This sub-problem would be the same as Eq. (\ref{solf}).
%\begin{equation}
%\max _{U^{T} U+V^{T} V=I} \operatorname{Tr}\left(U^{T} D_{U}^{-\frac{1}{2}} Z D_{V}^{-\frac{1}{2}} V\right)
%\end{equation}
%The problem can be solved easily by the  theorem\cite{nie2017learning},then we achieve F.
\subsection{Fix $F$, $Z$, Update $\lambda^v$ }
For simplicity, we denote $\|X^v-A^vZ^\top\|_F^2$ as $h^v$. Our objection function becomes %The second term can be omitted.Then the equation can be rewritten as follow:
\begin{equation}
H(\lambda^v)=\sum_{v=1}^{c} \lambda^{v} h^{v}+\sum_{v=1}^{c}\left(\lambda^{v}\right)^{\gamma}.
\end{equation}
It can be solved by taking the derivative on $\lambda^{(v)}$ and setting it to zero.
\begin{equation}
\frac{\partial H}{\partial \lambda^{v}}=h^{v}+\gamma\left(\lambda^{v}\right)^{\gamma-1}=0
\end{equation}
It yields 
\begin{equation}
\lambda^{v}=\left(-\frac{h^{v}}{\gamma}\right)^{\frac{1}{\gamma-1}}.
\label{soll}
\end{equation}
Because of $h^{v} \geq 0$ and $\gamma \textless 0$, $\lambda^{v}\geq 0$ is met. The complete steps to multi-view subspace clustering is outlined in Algorithm 2. It is worth pointing out that MSGL inherits all advantages of SGL, i.e., explicit graph structure, a linear complexity, extensions to out-of-sample data, convergence guarantee.  
\begin{algorithm}[H]%[tb]
\caption{MSGL for Subspace Clustering }
\label{alg:algorithm}
\textbf{Input}: Data matrix $[X^1; \cdots; X^c]$, anchor matrix $[A^1; \cdots; A^c]$, cluster number $k$,
 parameter $\alpha$, $\beta$, $\gamma$\\
\textbf{Output}: $k$ clusters
\begin{algorithmic}[1] %[1] enables line numbers

\STATE Initialize the matrix $F$ randomly and $\lambda^v$ as $1/c$.
%\STATE Apply k-means algorithm on each view of the data $X_i$ to obtain landmark matrix $A_i$
\WHILE{convergence condition does not meet}
\STATE Update $Z$ in Eq. (\ref{solmz}).
\STATE Update $F$ as Eq. (\ref{solf}).
\STATE Update $\lambda^v$ by Eq. (\ref{soll}).
\ENDWHILE
\STATE Run K-means on matrix $U$ to achieve final partition
\end{algorithmic}
\end{algorithm}
\begin{table}[H]			
			\renewcommand{\arraystretch}{1.1}
			\setlength{\tabcolsep}{8pt}{
			\begin{center}
			\caption{Description of the single-view data sets.\label{data}}
				\begin{tabular}{|c|cccc| }
					\hline
					 Data&Instance \#&Feature \#& Class \#&\\
					\hline	
BA&	1,404&	320&	120&	\\
% JAFFE&	213&	676&	10&	\\
ORL&	400&	1,024&	40&	\\
TR11&	414&	6,429&	9&	\\
TR41&	878&	7,454&	10&	\\
TR45&	690&	8,261&	10&	\\
RCV1-4&	9,625&	29,992&	4&	\\
%20Newsgroups&	18846&	26214&	20&	\\
% Reusters21578&	8293&	18933&	65&	\\
MNIST&	70,000&	784&	10&	\\
CoverType&	581,012&	54&	7&	\\
Pokerhand&	1,000,000&	10&	10&	\\
					\hline	
			\end{tabular}
	\end{center}		}
	\end{table}
\section{Single-view Experiments}\label{singleexp}
In this section, we conduct several experiments to evaluate our method on single-view data sets.
\subsection{Data Sets}
Nine popular data sets are tested. Specifically, BA\footnote{http://www.cs.nyu.edu/ roweis/data.html}, ORL\footnote{http://www.cl.cam.ac.uk/research/dtg/attarchive/facedatabase
.html}, MNIST\footnote{http://yann.lecun.com/exdb/mnist/} are image data, while TR11, TR41, TR4\footnote{http://www-users.cs.umn.edu/ han/data/tmdata.tar.gz}, RCV1-4 \cite{lewis2004rcv1} are text corpora. Pokerhand\footnote{https://archive.ics.uci.edu/ml/datasets/Poker+Hand} is evolutionary data and CoverType\footnote{http://archive.ics.uci.edu/ml/datasets/covertype} is collected to predict forest cover type based on cartographic variable. Table \ref{data} summarizes the statistics of these data sets. MNIST, CoverType, and Pokerhand are examined in out-of-sample problem.

\subsection{Comparison Methods}
For a fair comparison, we select five representative scalable clustering methods.

\textbf{ALRR}: an accelerated low-rank representation algorithm published in 2018 \cite{fan2018accelerated}.

\textbf{KMM}: a K-Multiple-Means method based on the partition of a bipartite graph published in 2019 \cite{nie2019k}.

\textbf{ESSC}: an efficient sparse subspace clustering algorithm published in 2020 \cite{pourkamali2020efficient}. %It proximal gradient method to search sparse representation vectors of data points that lie in a union of affine subspaces.

\textbf{FNC}: a directly solving normalized cut method published in 2018 \cite{chen2018spectral}. It has lower computational complexity and is fast for large-scale data.

\textbf{SSCOMP}: a popular sparse subspace clustering method based on orthogonal matching pursuit published in 2016 \cite{You2016Scalable}.

We tune the parameters in these methods to achieve the best performance. For our approach, clustering performance varies depending on the initialization of the K-means. Thus, we run 20 times and use a different seed to initialize K-means at each time. We report the mean and standard deviation values. Widely used clustering accuracy (ACC), normalized mutual information (NMI), and Purity are employed to evaluate the clustering performance \cite{chen2011large}. We conduct all experiments on a computer with a 2.6GHz Intel Xeon CPU and 64GB RAM, Matlab R2016a. The source code of our method is publicly available \footnote{https://github.com/sckangz/SGL}.
\subsection{Results}
Table \ref{res1} shows the clustering results of various methods. It can be seen that our proposed SGL outperforms other state-of-the-art techniques in most cases. In particular, our method performs much better than the most two recent methods KMM and ESSC. In terms of ACC, NMI, and Purity, SGL improves KMM by 12.57\%, 15.06\%, 14.49\% in average, respectively. With respect to ESSC, our gain is 16.66\%, 16.19\%, 12.17\%, respectively. Among the five competitive methods, ALRR gives more stable performance than others; SSCOMP is quite unstable, which could be caused by the fact that its strong assumptions are often broke facing real-world data. FNC generates mediocre results. 

We also list the consumed time for each method in Table \ref{res1}. We can observe that our method achieves comparable efficiency on those small size data sets, including AR, ORL, TR11, TR41, TR45. Though KMM demonstrates high efficiency, it clustering performance is degraded. Sparse subspace clustering based methods, i.e., ESSC and SSCOMP, cost much more time than others on TR11, TR41 and TR45. For medium size data RCV1-4, ALRR and ESSC run a long time. For example, ESSC takes 4078 seconds to finish the RCV1-4 data, while our method just needs 98.86 seconds. This indicates that our method is efficient yet effective.

\begin{table}%[!htbp]
\begin{center}
\renewcommand{\arraystretch}{1.1}
	\setlength{\tabcolsep}{1.2pt}{
\caption{Clustering performance on six data sets. The best performance is highlighted. Time is measured in seconds. \label{res1}}
%\scalebox{0.9}{
\begin{tabular}{|c|c|c|c|c|c|c|c|}
%\Xhline{1.0pt}
\hline
{Data}&{Metric} & ALRR& KMM& ESSC& FNC&SSCOMP&SGL\\
\hline

\multirow{4}*{BA}& ACC&39.03&41.45&28.56&46.79&18.66&\textbf{48.51(0.52)}\\
				
\cline{2-8}
&NMI&57.21&55.24&40.09&60.13&30.66&\textbf{60.47(0.22)}\\

\cline{2-8}
&PURITY&53.20&45.47&32.76&\textbf{60.13}&21.36&53.81(0.11)\\

\cline{2-8}
&TIME&8.07&	0.78&	18.30&	1.68&	0.31&	11.63(0.12)\\

\hline
\multirow{4}*{ORL}& ACC&68.00&
61.25&
60.75&
59.25&
63.75&
\textbf{68.83(0.39)}\\
				
\cline{2-8}
&NMI&
\textbf{82.73}&
77.16&
77.29&
77.33&
80.44&
81.90(0.15)\\

\cline{2-8}
&PURITY&
77.00&
65.00&
67.50&
64.25&
67.50&
\textbf{77.07(0.73)}\\

%\hline

\cline{2-8}
&TIME&	4.55
&	0.24
&	4.69
&	0.40
&	0.19
&3.23(0.15)\\

\hline

\multirow{4}*{TR11}& ACC&
72.94&
49.75&
52.90&
54.58&
64.97&
\textbf{75.31(0.86)}\\
				
\cline{2-8}
&NMI&
63.08&
27.94&
44.18&
41.35&
55.59&
\textbf{67.66(0.59)}\\

\cline{2-8}
&PURITY&
75.84&	
51.93&
57.73&
41.35&
78.01&
\textbf{80.32(1.67)}\\

%\hline

\cline{2-8}
&TIME
&	4.51
&	0.15
&	25.53
&	1.69
&	3.19
&5.31(0.08)
\\

\hline
\multirow{4}*{TR41}& ACC&
61.73&
66.05&
67.31&
58.42&
69.47&
\textbf{77.79(0)}\\
				
\cline{2-8}
&NMI&
62.87&
61.92&
65.57&
49.96&
66.73&
\textbf{72.11(0)}\\

\cline{2-8}
&PURITY&
73.00&
72.09&
73.35&
49.96&
80.52
&\textbf{84.16(0)}\\

%\hline

\cline{2-8}
&TIME&7.75
&0.36
&74.74
&6.83
&83.36
&7.63(0.23)\\

\hline
\multirow{4}*{TR45}& ACC&
72.31&
73.80&
67.25&
48.40&
73.04&
\textbf{74.41(2.44)}\\
				
\cline{2-8}
&NMI&
\textbf{70.10}&
67.73&
60.85&
31.22&
69.96&
69.53(1.82)\\

\cline{2-8}
&PURITY
&77.68&
82.34&
68.84&
31.22&
\textbf{83.76}
&
76.74(1.71)\\

%\hline

\cline{2-8}
&TIME&9.41
&0.35
&62.25
&4.83
&87.64
&6.34(0.18)\\

\hline
\multirow{4}*{RCV1-4}& ACC&
66.79&
47.63&
39.11&
65.37&
30.35&
\textbf{70.52(0)}\\
				
\cline{2-8}
&NMI&40.76&
17.12&
12.30&
35.00&
0.10&
\textbf{45.76(0)}\\

\cline{2-8}
&PURITY
&78.79&
47.63&
78.21&
35.00&
30.4&
\textbf{79.37(0)}
\\

%\hline

\cline{2-8}
&TIME&1175.80
&4.1
&4078.2
&85.90
&350.62
&98.86(0.67)\\

\hline

\end{tabular}}
			
	\end{center}
	
\end{table}

\subsection{Parameter Analysis}
\begin{figure*}[!htbp]
\centering
% \subfloat[ORL]{\includegraphics[width=.45\textwidth]{ORLanchor.pdf}}\\
% \subfloat[TR45]{\includegraphics[width=.45\textwidth]{TR45anchor.pdf}}\\
\includegraphics[width=.45\textwidth]{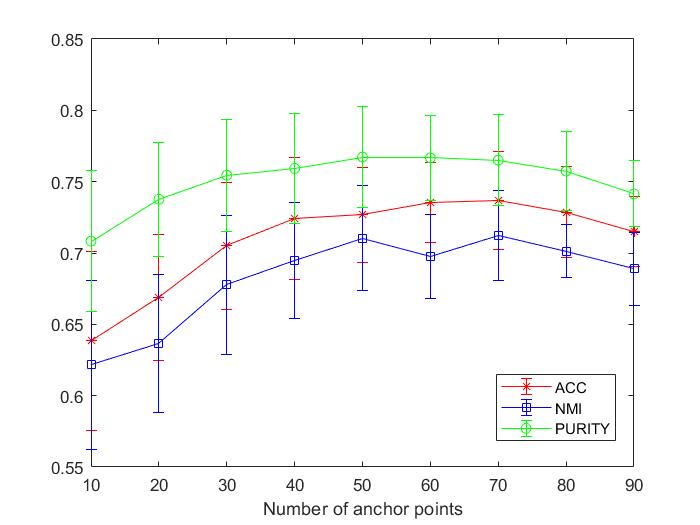}
\hspace{0.cm}
\includegraphics[width=.45\textwidth]{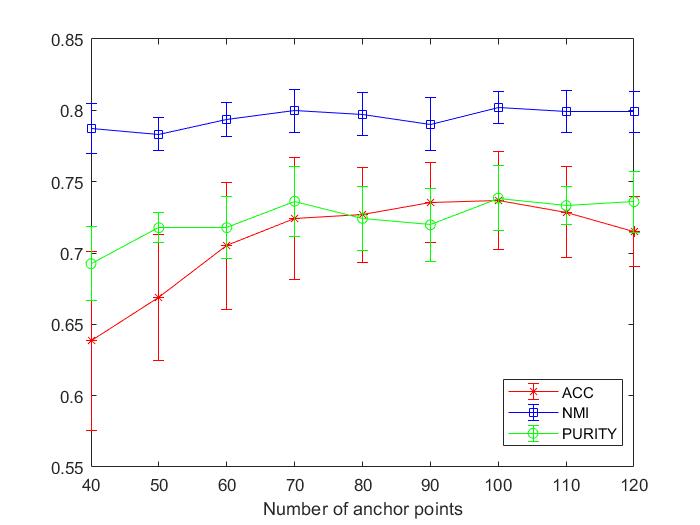}\\
% \includegraphics[width=.42\textwidth]{ESCTR45.jpg}
% \hspace{0.cm}
% \includegraphics[width=.42\textwidth]{ESCORL.jpg}
\caption{The impact of anchor number and initialization on TR45 (the left) and ORL (the right) data sets. The average performance and error bar are displayed. } \label{anchors}
\end{figure*}

There are three parameters $\alpha, \beta$ and anchor number $m$ in our model. We can observe that the dictionary A is constituted of anchor points, which is important to the proposed model. Recently, some advanced anchor points selection methods are developed. For example, \cite{abdolali2019scalable} proposes to select a few sets of anchors points based on randomized hierarchical clustering. Consequently, it selects a different set of anchors for each data point. \cite{you2020self} adopts an exemplar selection method FFS to find a subset that best reconstructs all data points based on the $\ell_1$ norm of the representation coefficients. Compared to them, our adopted K-means approach is simple and straightforward. However, K-means is sensitive to initialization. Therefore, we discuss the clustering effect caused by the variations of anchor number and initialization. It is reasonable to assume that we at least need $k$ anchors to reveal the underlying subspaces. In Fig. \ref{anchors}, we let the number of anchor points varies over the range $[40,\cdots,120]$ and $[10,\cdots,90]$ for ORL and TR45, respectively. It shows that the clustering performance indeed changes along with the variation of anchor number and initialization. However, it can be seen that we don't require too many anchors for a good performance. Furthermore, the standard variation suggests that our results change slightly due to the initialization. 
%Besides, we repeat the experiments twenty times and use a different seed to initialize K-means at each time. Then, we report the average value and standard deviation of the clustering results. 

\begin{figure*}[!htbp]
\centering
% \subfloat[ORL]{\includegraphics[width=.45\textwidth]{ORLanchor.pdf}}\\
% \subfloat[TR45]{\includegraphics[width=.45\textwidth]{TR45anchor.pdf}}\\
% \includegraphics[width=.42\textwidth]{ourtr45whole.jpg}
% \hspace{0.cm}
% \includegraphics[width=.42\textwidth]{ourorlwhole.jpg}\\
\includegraphics[width=.45\textwidth]{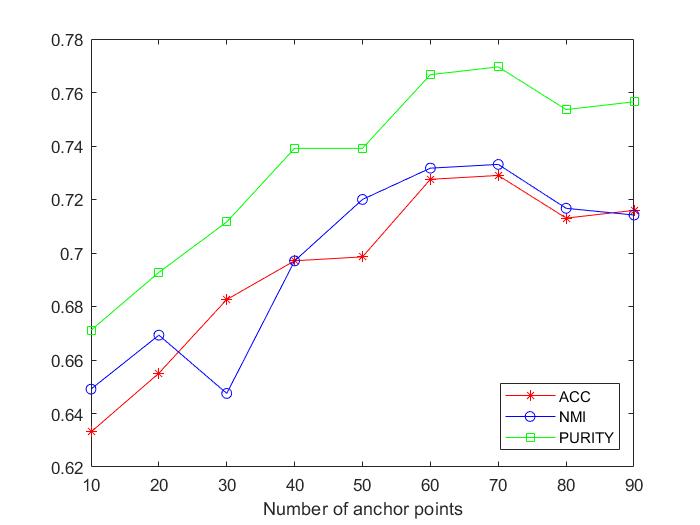}
\hspace{0.cm}
\includegraphics[width=.45\textwidth]{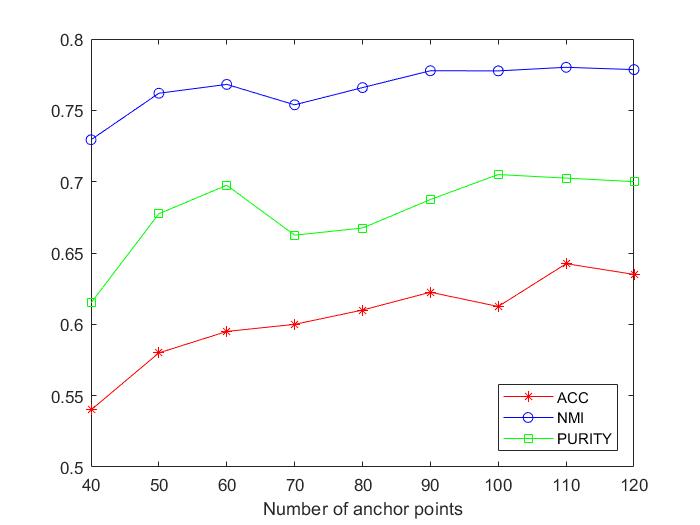}
\caption{The performance of SGL on TR45 (the left) and ORL (the right) data sets, where anchors are chosen by FFS. } \label{anchors1}
\end{figure*}

Furthermore, we adopt FFS \cite{you2020self} to select anchors and use them to build an anchor matrix $A$. Then we plot the performance of our method in Fig. \ref{anchors1}. We can see that its performance is inferior to K-means based approach. This could be caused by the fact that these anchors are close to a few cluster centroids, thus K-means is more appropriate than FFS. FFS could well handles the case when anchors lie close to a union of subspaces \cite{you2020self}.
%our method is very stable when the number of anchor points is greater than 2*$k$($k$ is the number of our clusters).Our method is not sensitive to the value of the number of anchor points.

\begin{figure*}[!htbp]
\centering
\includegraphics[width=.31\textwidth]{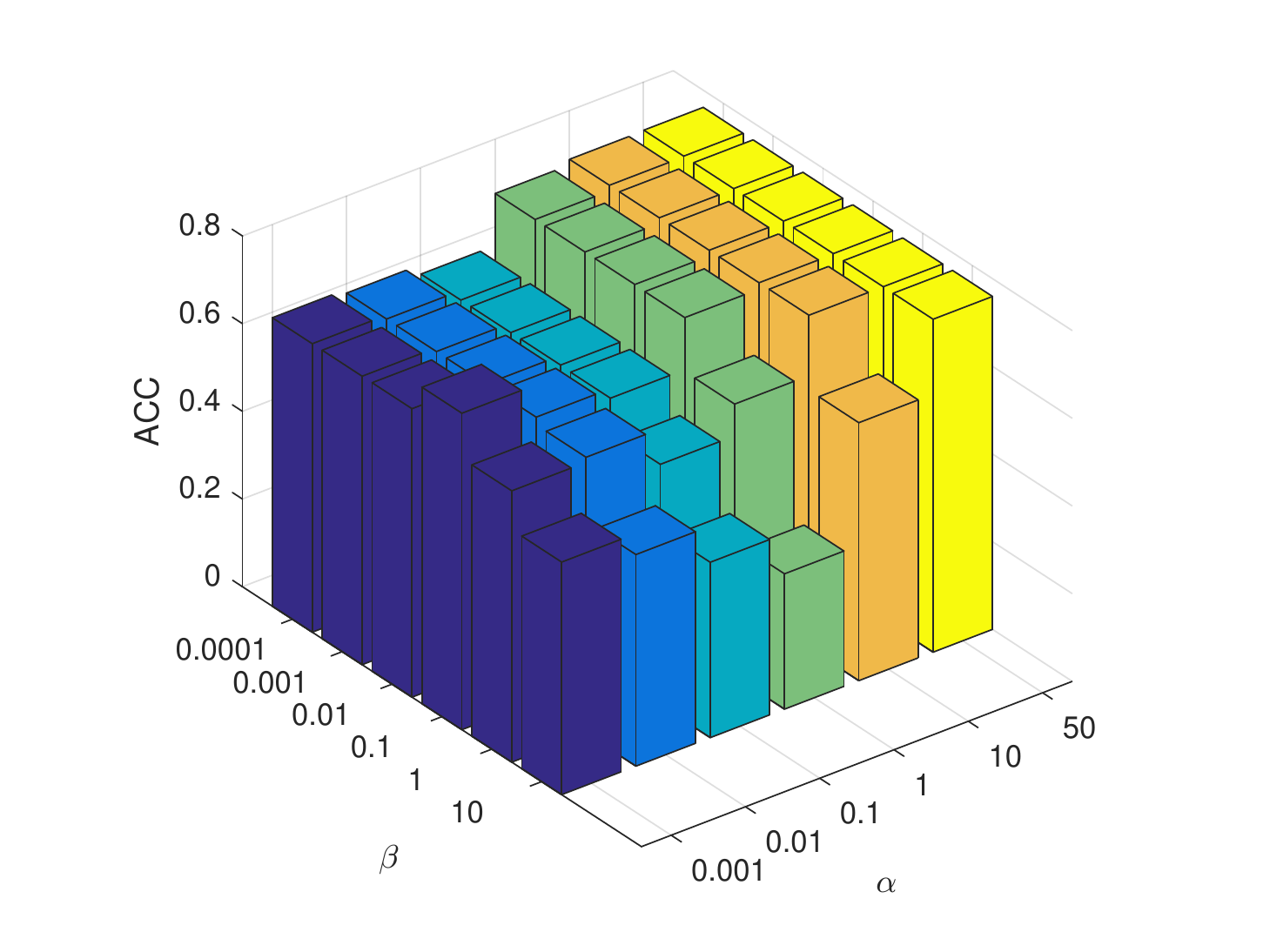}
\hspace{0.cm}
\includegraphics[width=.31\textwidth]{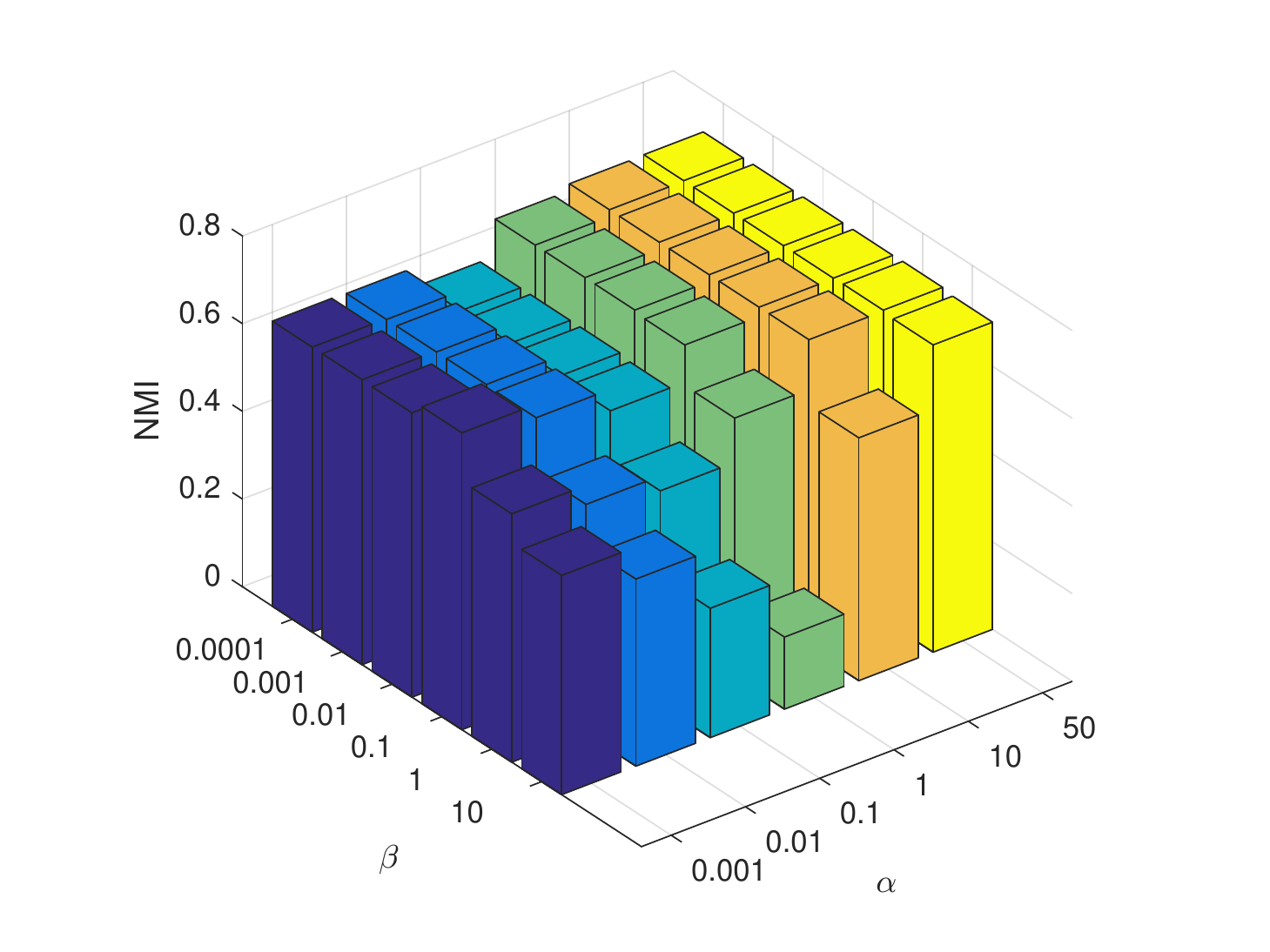}
\hspace{0.cm}
\includegraphics[width=.31\textwidth]{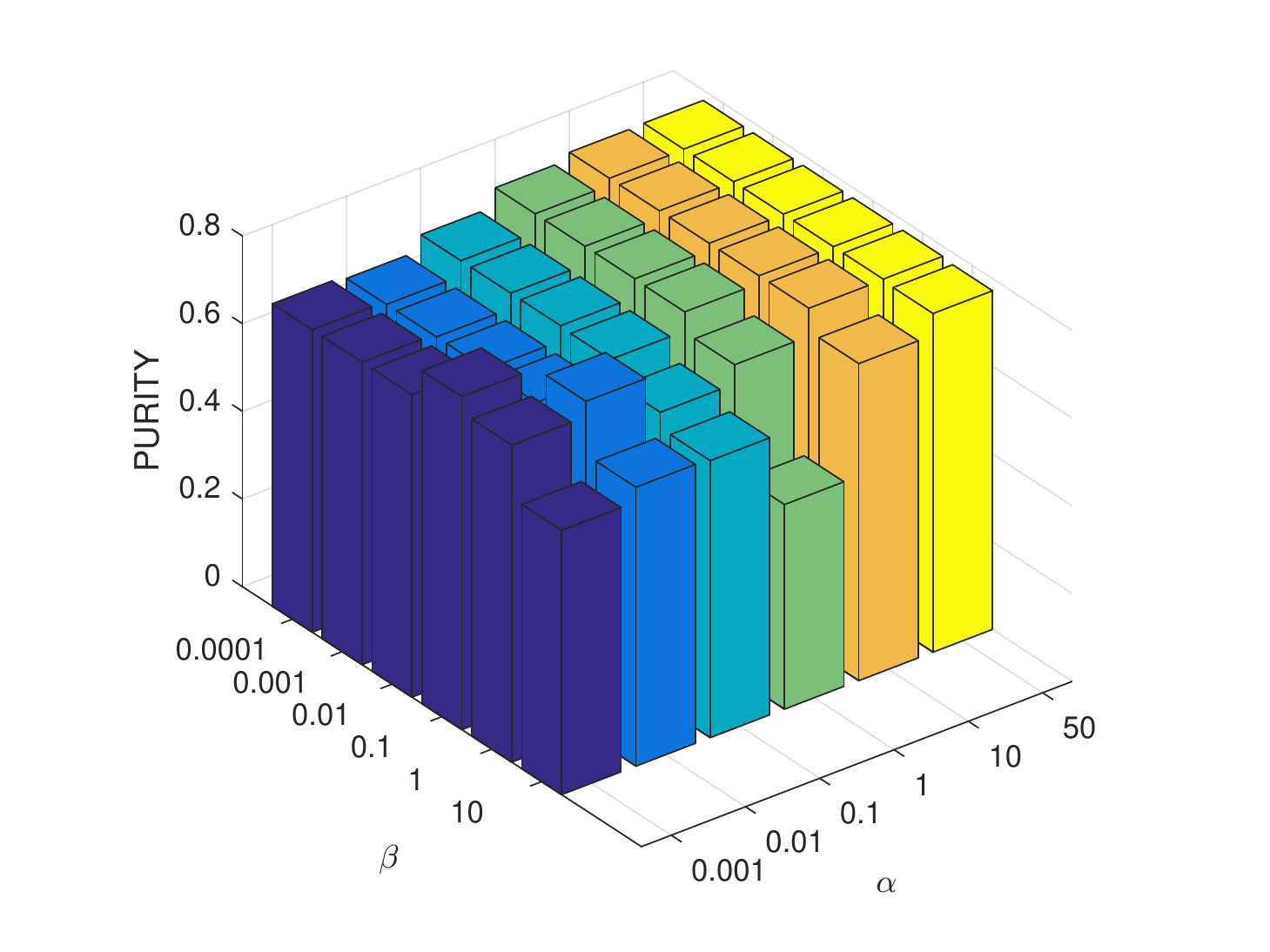}\\
\includegraphics[width=.31\textwidth]{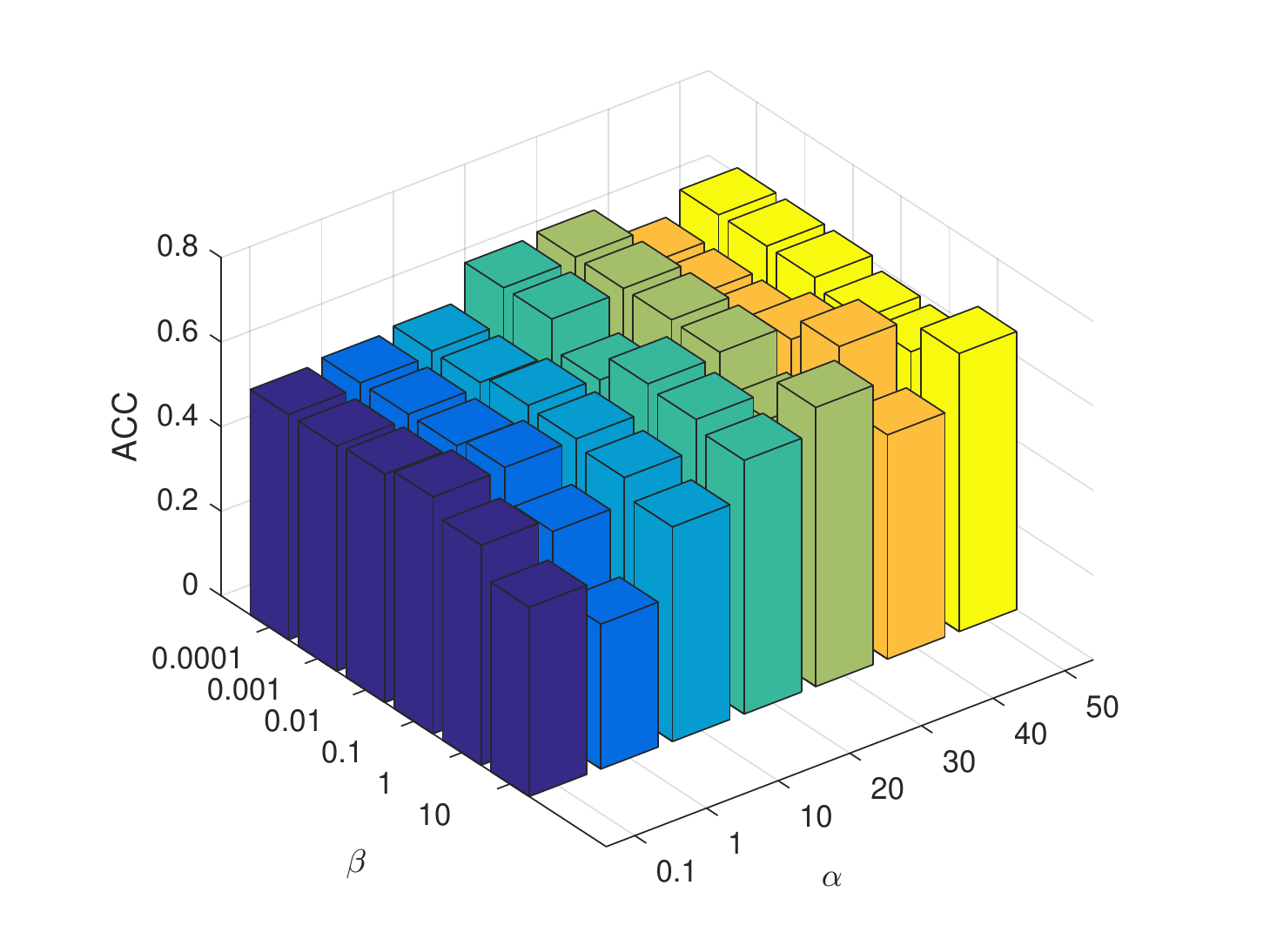}
\hspace{0.cm}
\includegraphics[width=.31\textwidth]{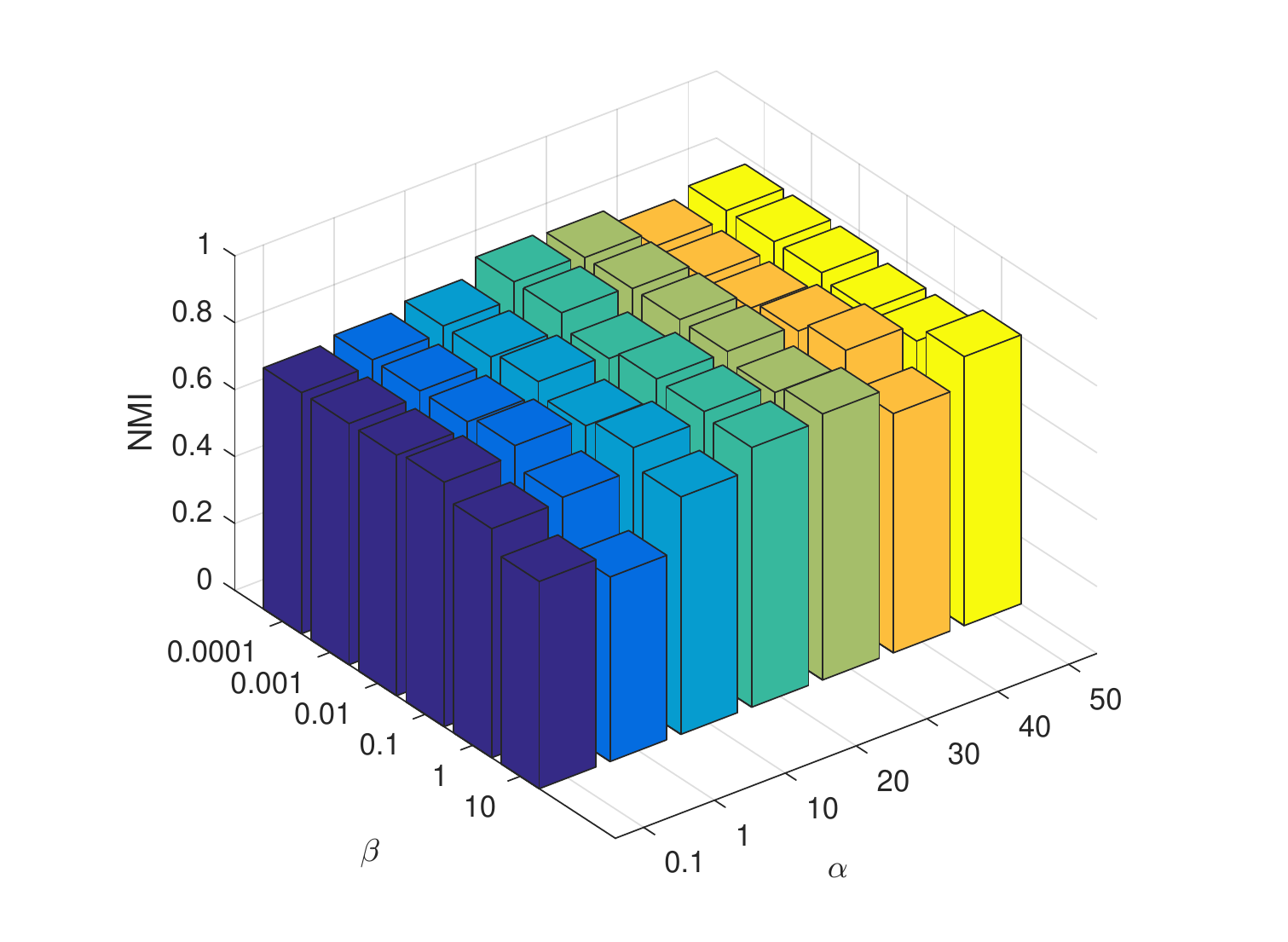}
\hspace{0.cm}
\includegraphics[width=.31\textwidth]{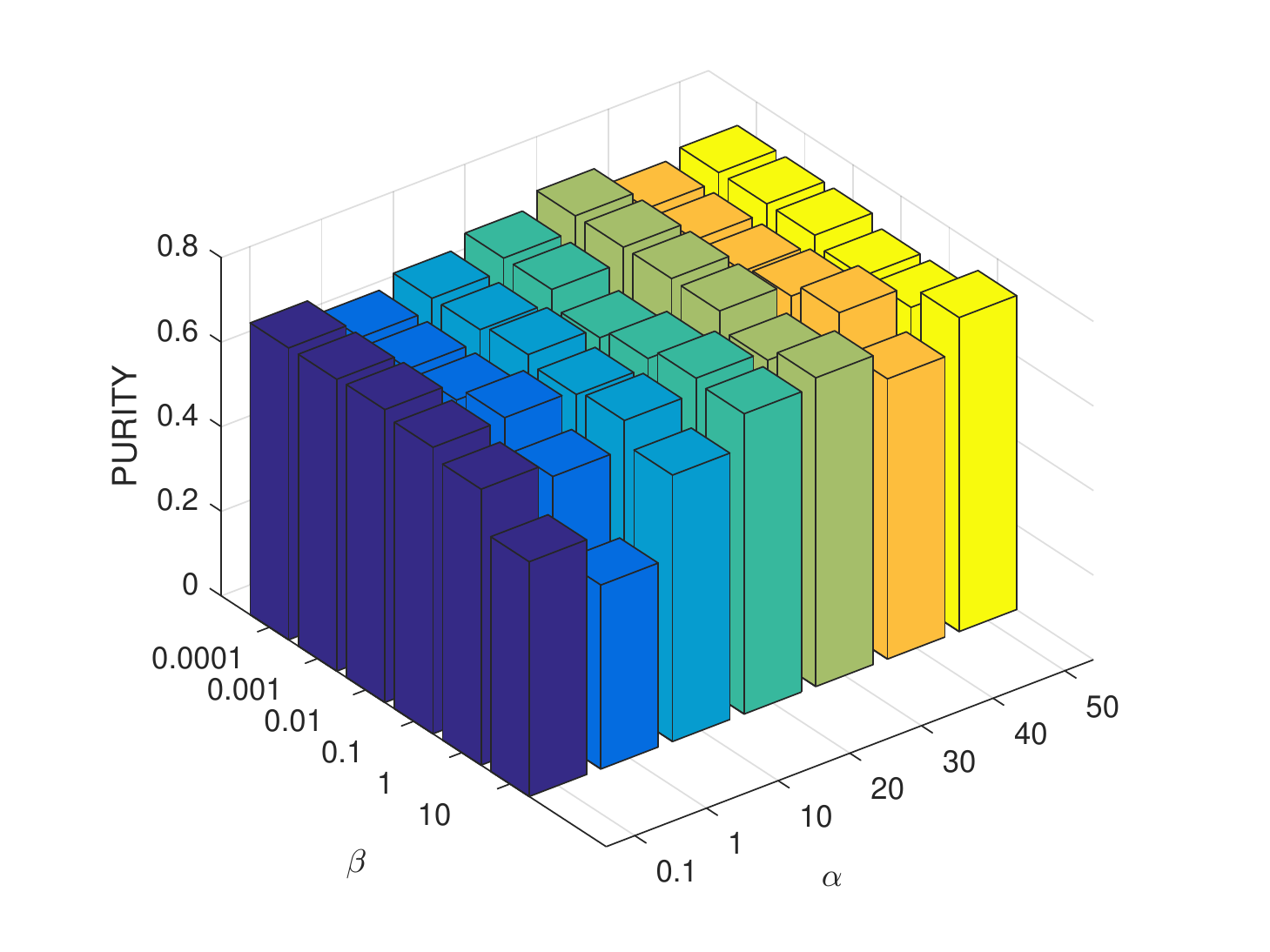}
\caption{Sensitivity demonstration of parameters $\alpha$ and $\beta$ for our method over TR45 (the 1st row) and ORL (the 2nd row) data sets.} \label{para}
\end{figure*}

Next, we fix the number of anchor points and analyze the sensitive of $\alpha$ and $\beta$ on TR45 and ORL data sets. $\beta$ varies over the range of $[0.0001,0.001,0.01,0.1,1,10]$. $\alpha$ varies over $[0.001,0.01,0.1,1,10,50]$ and $[0.1,1,10,20,30,40,50]$ on TR45 and ORL respectively. Fig. \ref{para} displays how the clustering results of our method vary with $\alpha$ and $\beta$. We can find that the performance of our method is very stable with respect to a large range of $\alpha$ and $\beta$ values. In practice, we can fix $\alpha$ and tune $\beta$.

\subsection{Out-of-Sample Experiment}
In this subsection, we conduct experiments on MNIST, CoverType and Pokerhand to evaluate our method for addressing out-of-sample problem. Following the setting in SLSR \cite{peng2015unified}, we randomly select 1000 data points as in-sample data for CoverType and Pokerhand, while 2000 data points for MNIST. The rest data are regarded as out-of-sample data for testing. As described in subsection \ref{outof}, we apply kNN algorithm to test out-of-sample data. Concretely, we adopt 3NN and 1NN to the anchor points and their labels to predict the clusters of test data. As a baseline, we also apply 1NN to the in-sample data. Moreover, we compare with SLSR \cite{peng2015unified}, to our best knowledge, which is the only method proposed to address out-of-sample problem for subspace clustering. %\textcolor{red}{We run 20 times on our method and report the average values and the standard deviation.}%Then we compare its ACC, NMI, and time .In order to demonstrate the extension of our method to the out-of-sample data,we choose  as the baseline method.Besides,according to our method,

We report the results in Table \ref{outres}. As for the accuracy, we can find that our approach consistently outperforms other methods. In particular, our accuracy surpasses SLSR method by about 10\% on CoverType and Pokerhand data. Our method obtains comparable performance with baseline on NMI and Purity. These results suggest that our anchor points can well represent the structure of the raw data. We can also see that the number of neighbors in kNN has a big influence to the final performance. This confirms the previous observation in the literature. 
 
In terms of running time, we can observe that our approach outperforms others by a large margin and can finish 1M samples in less than 1 second. Compared to SLSR, our method is at least 200 times faster. With respect to conventional kNN approach, our method also runs much faster since we use fewer points. In summary, our method is a promising approach to process out-of-sample problem. On the other hand, this approach demonstrates itself
to be an alternative way to tackle big data challenge.

% \begin{table}[t]%[!htbp]
% \begin{center}
% \renewcommand{\arraystretch}{3.1}
% 	\setlength{\tabcolsep}{1.5pt}{
% \caption{Clustering performance on out-of-sample problem.\label{mnist}}
% %\scalebox{0.9}{
% \begin{tabular}{|c|c|c|c|c|c|}
% %\Xhline{1.0pt}
% \hline
% {data set}&{Method} & Acc& NMI& Purity& Time (s)\\
% \hline

% \multirow{2}*{MNIST(1/7)}&1nn+in-sample data&61.88&58.55&67.13&98.2\\
				
% \cline{2-6}
% &1nn+anchor points&62.18&58.27&66.63&0.84\\

% \hline

% \multirow{2}*{MNIST(3/7)}&1nn+in-sample data&64.09&59.92&68.89&198.63\\
				
% \cline{2-6}
% &1nn+anchor points&64.13&59.67&68.53&0.93\\

% \hline

% \multirow{2}*{Covtype(1/7)}&1nn+in-sample data&32.25&11.22&53.56&487.5\\
% \cline{2-6}
% &1nn+anchor points&31.83&7.73&50.53&0.77\\

% \hline

% \multirow{2}*{Covtype(3/7)}&1nn+in-sample data&30.93&10.90&54.31&984.29\\
% \cline{2-6}
% &1nn+anchor points&29.85&9.49&50.24&0.86\\

% \hline
% \multirow{2}*{RCV1(1/7)}&1nn+in-sample data&	14.23&20.10&24.73&26087\\
% \cline{2-6}
% &1nn+anchor points&	13.09&12.40&18.57&1716.91\\

% \hline

% \multirow{2}*{RCV1(3/7)}&1nn+in-sample data&	16.92&23.59&27.16&50621\\
% \cline{2-6}
% &1nn+anchor points&	17.13&21.18&25.56&1760\\

% \hline
% \end{tabular}}
			
% 	\end{center}
	
% \end{table}
\begin{table}[t]%[!htbp]
\begin{center}
\renewcommand{\arraystretch}{1.2}
	\setlength{\tabcolsep}{1.5pt}{
\caption{Clustering performance on out-of-sample problem.\label{outres}}
%\scalebox{0.9}{
\begin{tabular}{|c|c|c|c|c|c|}
%\Xhline{1.0pt}
\hline
{Data}&{Method} & Acc& NMI&PURITY& Time (s)\\
\hline

\multirow{4}*{MNIST}&3NN+anchor points&57.23(1.31)&53.72(0.84)&60.57(0.64)&{1.04(0.02)}\\
				
\cline{2-6}
&1NN+anchor points&\textbf{58.23(0.35)}&\textbf{54.26(0.24)}&\textbf{62.37(0.19)}&0.67(0.01)\\

\cline{2-6}
&1NN+in-sample data&56.49(0.75)&51.91(0.79)&59.61(0.96)&22.69(0.66)\\

\cline{2-6}
&SLSR&52.26&47.72&57.06&252.28\\
\hline

\multirow{4}*{CoverType}&3NN+anchor points&\textbf{39.01(0.92)}&\textbf{8.57(0.78)}&50.15(0.41)&3.05(0.04)\\
				
\cline{2-6}
&1NN+anchor points&32.10(1.48)&6.69(0.52)&50.21(0.13)&0.64(0.01)\\

\cline{2-6}
&1NN+in-sample data&31.41(1.21)&6.47(0.57)&\textbf{52.36(0.28)}&8.24(0.11)\\

\cline{2-6}
&SLSR&26.5&7.3&49.42&178.72\\
\hline

\multirow{4}*{Pokerhand}&3NN+anchor points&\textbf{27.97(1.71)}&0.27(0.10)&50.73(0.45)&8.81(0.07)\\
\cline{2-6}
&1NN+anchor points&22.15(1.48)&\textbf{0.33(0.10)}&\textbf{50.90(0.47)}&0.88(0.01)\\

\cline{2-6}
&1NN+in-sample data&19.73(1.36)&0.17(0.10)&50.23(0.38)&2.34(0.12)\\

\cline{2-6}
&SLSR&15.82&0.06&50.20&284.52\\
\hline
\end{tabular}}
			
	\end{center}
	
\end{table}

 \begin{figure*}[!htbp]
\centering
\includegraphics[width=0.31\textwidth]{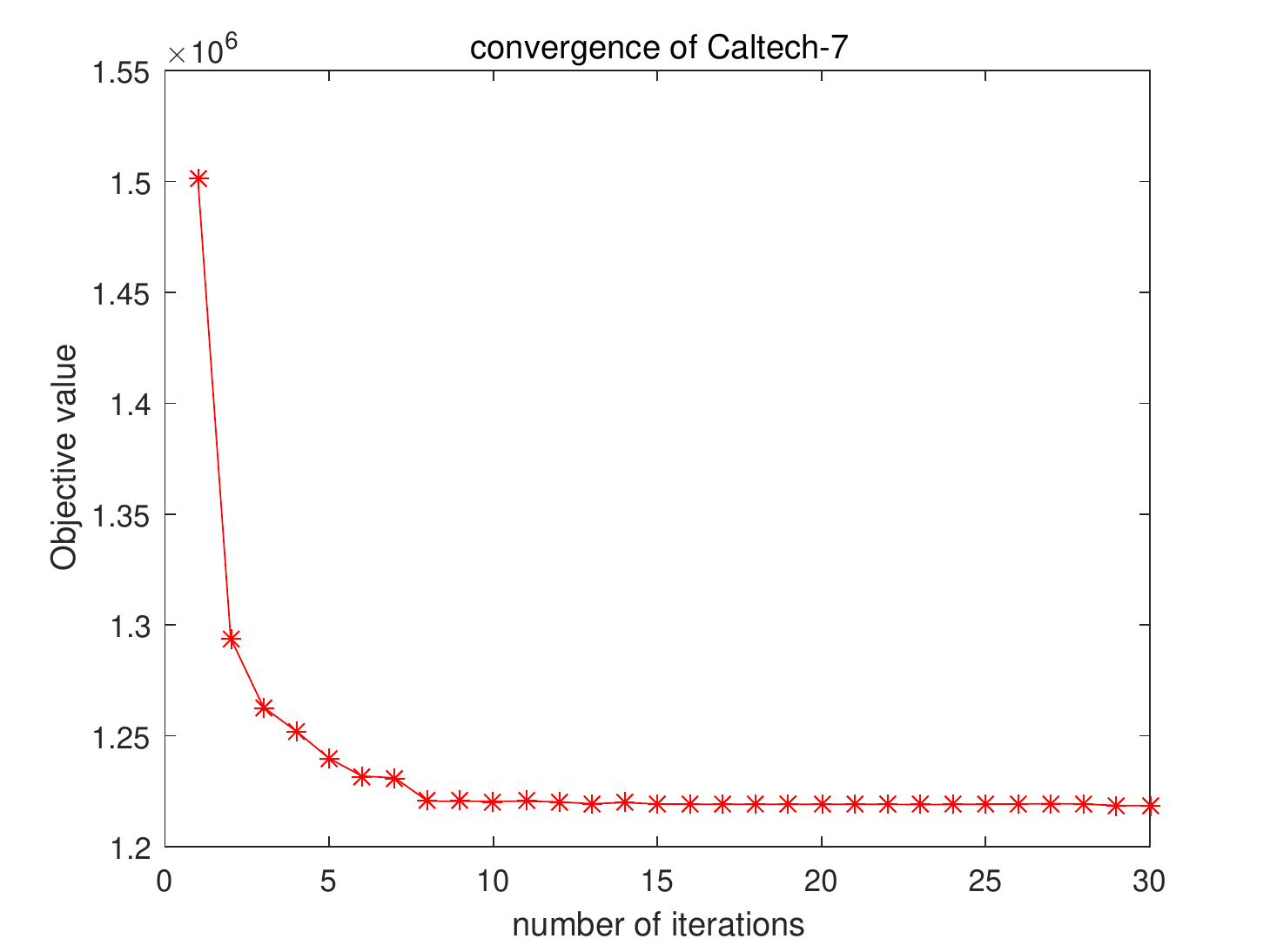}
\hspace{0.cm}
\includegraphics[width=.31\textwidth]{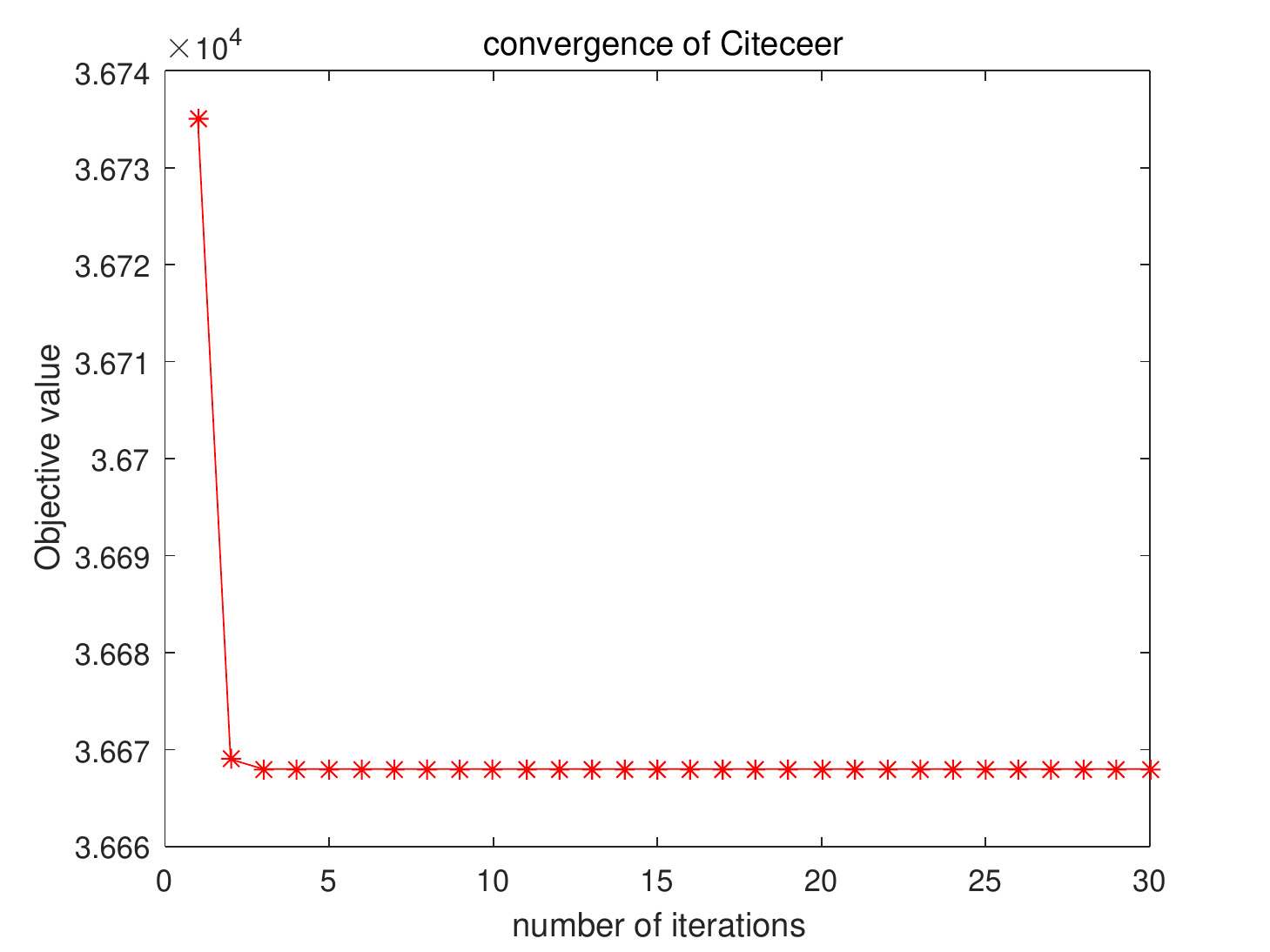}
\includegraphics[width=.31\textwidth]{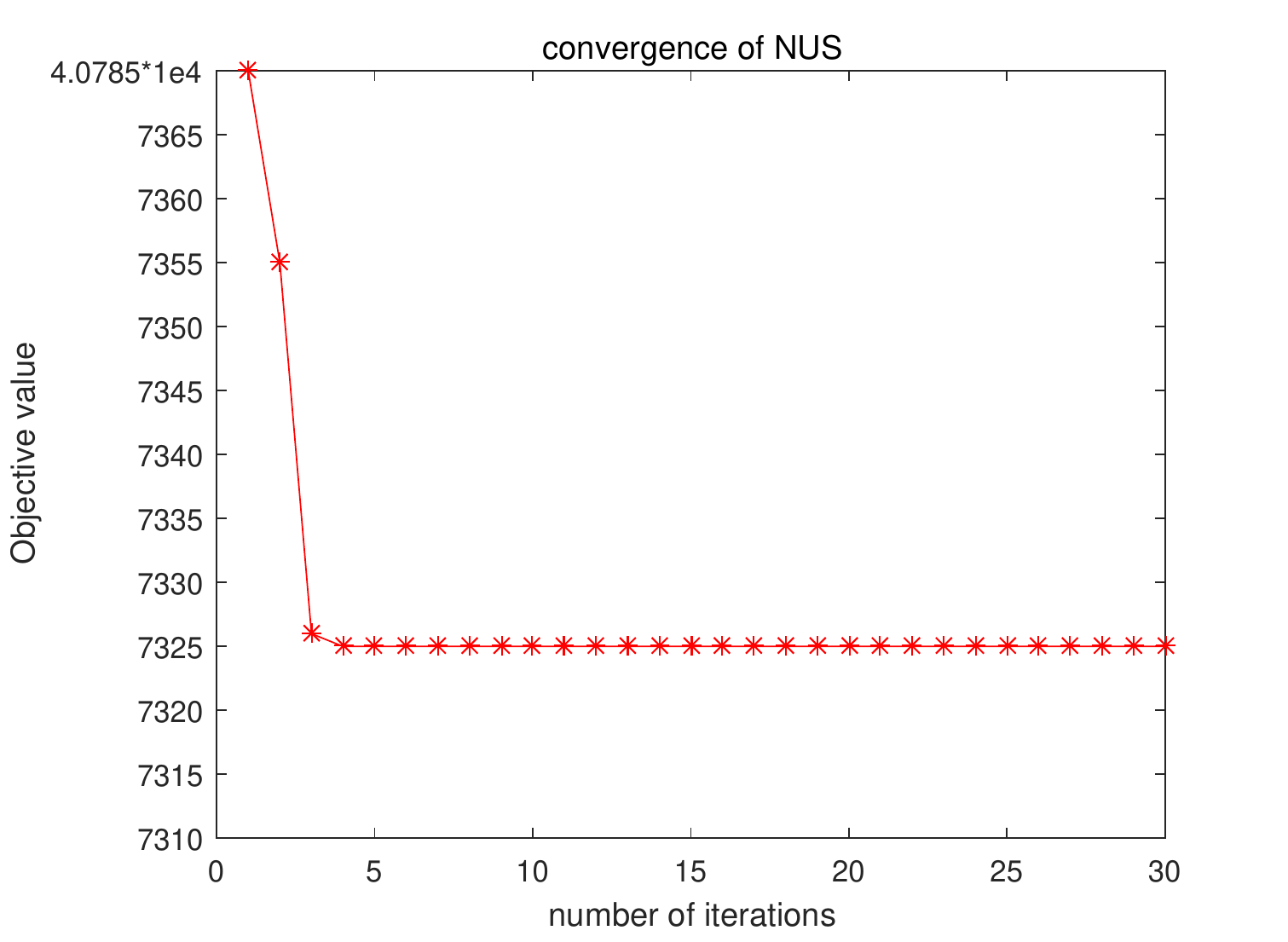}

\caption{The evolution of objective function value of (\ref{multiview}) on all multi-view data sets. } \label{converge}
\end{figure*}
\section{Multi-view Experiments}\label{multiexp}
In this section, we assess the effectiveness and efficiency of our multi-view model (\ref{multiview}). 
\subsection{Setup}
We perform experiments on three benchmark data sets: Caltech-7\footnote{http://www.vision.caltech.edu/ImageDatasets/Caltech101/}, NUS\footnote{https://lms.comp.nus.edu.sg/wp-content/uploads/2019/research/nuswide/NUS-WIDE.html}, Citeseer\footnote{http://lig-membres.imag.fr/grimal/data.html}.
Both Caltech-7 and NUS are object recognition database, while Citeseer is a document data with content and citations. The details of them are summarized in Table \ref{multiviewdata}.
\begin{table}[!hbtp]
\begin{center}
\renewcommand{\arraystretch}{1.}
\caption{Statistical information of the multi-view data sets. The number in parenthesis denotes dimension. \label{multiviewdata}}
\label{multidata} \scalebox{.85}{
\begin{tabular}{llll}
\hline%{1.0pt}%
%\hline

{View} & {Caltech-7}  & {Citeseer} & {NUS}  \\\hline
1& Gabor (48) & Content (3703) & Color Histogram (65)\\
2&  Wavelet moments (40)& Citation (3312)& Color moments (226)\\
3& CENTRIST (254)&-& Color correlation (145)\\
4 &  HOG (1984)& -& Edge distribution (74)\\
5&  GIST (512)&-& Wavelet texture (129) \\
6& LBP (928)& - &- \\\hline
Data points &  1474&3312 & 30000\\
Class number& 7 & 6 & 31\\
%Type & image  & image & image \\
%\Xhline{1.0pt}%
\hline
\end{tabular}}
\end{center}
\end{table}
We compare our proposed MSGL method with four other state-of-the-art multi-view methods.\\
\textbf{AMGL} \cite{nie2016parameter} is a popular multi-view spectral clustering method proposed in 2016. Though it is parameter-free, it has a high complexity since it involves SVD implemented on $n\times n$ matrix in each iteration. \\
\textbf{MLRSSC} \cite{brbic2018} is a multi-view subspace clustering method developed in 2018. It has good performance since it combines low-rank and sparse model, but it has $\mathcal{O}(n^3)$ complexity.\\
\textbf{MSC\_IAS} \cite{wang2019multi} is proposed to learn a better graph in latent space in 2019. It surpasses a number of multi-view subspace clustering methods, but it also has $\mathcal{O}(n^3)$ complexity.\\
\textbf{LMVSC} \cite{kang2019large} is a linear multi-view subspace clustering method published in 2020. It shows superior performance and high efficiency.

We use grid search to find the best parameters for all methods. For MSGL, $\gamma$ is searched from $[-1,-2,-3,-4,-5]$. %We also run 20 times on our method  and report the average values and the standard deviation.}
\subsection{Results}
Table \ref{cal}-\ref{nus} report the clustering performance on the three data sets. Both AMGL and MLRSSC raise out-of-memory exception on the NUS data. We can observe that our method outperforms other methods in most cases, including our closest competitor LMVSC. In particular, our method MSGL constantly outperforms LMVSC on accuracy and Purity. In terms of NMI, our method also achieves comparable or even better performance than LMVSC. AMGL, MLRSSC and MSC\_IAS produce poor accuracy and NMI on Caltech-7 and Citeseer data sets. There are some possible causes for this. AMGL uses losses of different views to weight each view, which might not be flexible to distinguish the contributions of diverse views. Moreover, some real-world data sets can not be simply characterized by low-rank or sparse structure as used in MLRSSC and MSC\_IAS. By contrast, we directly consider the cluster structure, which is a target-oriented solution.

For running time comparison, the proposed method is usually faster than the baseline methods except LMVSC. Though both MSGL and LMVSC have linear time complexity, LMVSC is faster since it is iteration-free. To be precise, LMVSC does not consider the structure of graph and distinguish views, so it is a one-pass approach. For large-scale
 data NUS, MSC\_IAS costs about 12 hours, while our method can finish it in 10 minutes. Hence, in term of efficiency and effectiveness, MSGL and LMVSC are appealing in practice applications.
\begin{table}[H]			
	\renewcommand{\arraystretch}{0.99}
	
	\setlength{\tabcolsep}{6pt}{
		\begin{center}
			\caption{Clustering performance on Caltech-7 data.
\label{cal}}
			\begin{tabular}{|c |c | c|c| c|  }
				\hline
				Method& ACC& NMI& PURITY& Time (s)\\
				\hline	
				AMGL&	45.18&	42.43&	46.74&	20.12\\
                MLRSSC&	37.31&	21.11&	41.45&	22.26\\
                MSC\_IAS&39.76&24.55&44.44&57.18\\
                LMVSC&	72.66&	51.93&	75.17&	135.79\\
                MSGL& \textbf{73.31(0.96)}& \textbf{52.47(2.23)}&  \textbf{77.34(2.87)}& 395.63(21.33)\\
				\hline			
			\end{tabular}
			
	\end{center}		}
	
\end{table}

\begin{table}[H]			
	\renewcommand{\arraystretch}{0.99}
	
	\setlength{\tabcolsep}{6pt}{
		\begin{center}
			\caption{Clustering performance on Citeseer data.
\label{hw}}
			\begin{tabular}{|c |c | c|c| c| }
				\hline
				Method& ACC& NMI& PURITY& Time (s)\\
				\hline	
				AMGL&	16.87&	0.23&	16.87&	449.07\\
                MLRSSC&	25.09&	02.67&	63.70&	106.1\\
                MSC\_IAS&34.11&11.53&\textbf{80.76}&191.29\\
                LMVSC&	52.26&	25.71&	54.46&	21.33\\
                MSGL& \textbf{54.47(0.78)}& \textbf{26.54(0.94)}&  57.49(1.13)& 62.23(4.72)\\
				\hline			
			\end{tabular}
			
	\end{center}		}
	
\end{table}

\begin{table}[H]			
	\renewcommand{\arraystretch}{0.99}
	
	\setlength{\tabcolsep}{6pt}{
		\begin{center}
			\caption{Clustering performance on NUS data.
\label{nus}}
			\begin{tabular}{|c |c | c|c| c| }
				\hline
				Method& ACC& NMI& PURITY& Time (s)\\
				\hline	
				MSC\_IAS& 15.48& \textbf{15.21}& 16.75& 45386\\
			    LMVSC&15.53&	12.95&	19.82&	165.39\\
			MSGL &\textbf{16.31(0.42)}& 12.26(0.28)& \textbf{20.51(0.37)}& 547.58(32.11)\\
				\hline			
			\end{tabular}
			
	\end{center}		}
\end{table}
\subsection{Convergence Analysis}
As mentioned earlier, MSGL is a convergent algorithm. To verify this, we demonstrate the behavior of the objective value of Eq. (\ref{multiview}) in Fig. \ref{converge}. It shows that our algorithm converges within 10 iterations on all three
real-world data sets. Once again, this supports that our method is efficient.

%For example, as for the citeseer and NUS data, we can converge in five iterations. Caltech-7 can finish in 10 iterations. All these results can prove the efficiency of our method. Through our method, we learn the representation of the raw data very fast, thus making our objective function vary a little when the number of iterations are increasing. 
\section{Conclusion}\label{conclude}
In this paper, we propose a novel graph-based subspace clustering framework to cope with single view and multi-view data. We simultaneous consider graph structure, scalability, and out-of-sample problems by making use of the anchor idea, bipartite graph and spectral graph property. Consequently, a graph with explicit cluster structure is learned in linear complexity. Theoretical analysis builds the connection between our method and K-means clustering. Extensive experimental results demonstrate that our method can reduce the time complexity without sacrificing the clustering performance. In the future, we plan to investigate new anchor selection strategy to improve the stableness of the proposed approach.

\bibliographystyle{IEEEtran}

\bibliography{ref,ref2}

\end{document}